\pdfoutput=1
\documentclass[nohyperref]{article}

\usepackage[utf8]{inputenc}

\usepackage{microtype}
\usepackage{graphicx}
\usepackage{subfigure}
\usepackage{booktabs} 
\usepackage{xurl}

\usepackage{tikz}
\usetikzlibrary{matrix}
\usepackage{makecell}
\usepackage{xcolor}
\usepackage{tcolorbox}
\usepackage{bm}
\usepackage{multirow}
\usepackage{pgfplots}
\usepackage{arydshln}
\usepackage{array}
\usepackage{xspace}
\usepackage{wrapfig}
\usepackage{hyperref}
\usepackage{pifont}
 \usepackage{ulem}
\usepackage[algo2e,ruled,linesnumbered]{algorithm2e}

\definecolor{tiffanyblue}{RGB}{129,216,208}
\definecolor{bangdiblue}{RGB}{0,149,182}
\definecolor{kleinblue}{RGB}{0,47,167}
\definecolor{kabuliblue}{RGB}{26,85,153}
\definecolor{purple}{RGB}{138,43,226}

\usepackage[accepted]{icml2022}

\usepackage{amsmath}
\usepackage{amssymb}
\usepackage{mathtools}
\usepackage{amsthm}

\usepackage[capitalize,noabbrev]{cleveref}
\crefname{section}{§}{§§}
\Crefname{section}{§}{§§}

\theoremstyle{plain}

\theoremstyle{definition}

\theoremstyle{remark}

\usepackage[textsize=tiny]{todonotes}

\icmltitlerunning{Learning Multiscale Transformer Models for Sequence Generation}

\begin{document}

\twocolumn[
\icmltitle{Learning Multiscale Transformer Models for Sequence Generation}



\icmlsetsymbol{equal}{*}

\begin{icmlauthorlist}
\icmlauthor{Bei Li}{equal,neu}
\icmlauthor{Tong Zheng}{equal,neu}
\icmlauthor{Yi Jing}{equal,neu}
\icmlauthor{Chengbo Jiao}{niu}
\icmlauthor{Tong Xiao}{neu,niu}
\icmlauthor{Jingbo Zhu}{neu,niu}
\end{icmlauthorlist}

\icmlaffiliation{neu}{School of Computer Science and Engineering, Northeastern University, Shenyang, China}
\icmlaffiliation{niu}{NiuTrans Research, Shenyang, China}

\icmlcorrespondingauthor{Tong Xiao}{xiaotong@mail.neu.edu.cn}

\icmlkeywords{Machine Learning, ICML}

\vskip 0.3in
]



\printAffiliationsAndNotice{\icmlEqualContribution} 

\begin{abstract}

    Multiscale feature hierarchies have been witnessed the success in the computer vision area. This further motivates researchers to design multiscale Transformer for natural language processing, mostly based on the self-attention mechanism. For example, restricting the receptive field across heads or extracting local fine-grained features via convolutions. However, most of existing works directly modeled local features but ignored the word-boundary information. This results in redundant and ambiguous attention distributions, which lacks of interpretability. In this work, we define those scales in different linguistic units, including sub-words, words and phrases. We built a multiscale Transformer model by establishing relationships among scales based on word-boundary information and phrase-level prior knowledge. The proposed \textbf{U}niversal \textbf{M}ulti\textbf{S}cale \textbf{T}ransformer, namely \textsc{Umst}, was evaluated on two sequence generation tasks. Notably, it yielded consistent performance gains over the strong baseline on several test sets without sacrificing the efficiency.
    \end{abstract}

    \section{Introduction}
    \label{sec:intro}
    Transformers \cite{vaswani2017attention} have achieved remarkable success on a wide range of tasks in natural language processing (NLP) \cite{devlin-etal-2019-bert,dai-etal-2019-transformer,wang-etal-2019-learning,Dehghani2019universal}. Given an input sequence, it uses self-attention mechanisms to establish global interactions over all positions without recurrences or convolutions.

    Despite great potential on most of NLP tasks, the Transformer backbones still have a major shortcoming that it does not model local fine-grained features explicitly. For example, given a word in a sentence, the surrounding words have more potential correlations with it than distant words. Also, replacing the self-attention mechanism by dynamic convolutions even shows competitive favorable results with Transformer \cite{wu2019DynamicConv}. This observation motivates researchers to combine local and global features for a multiscale representation of features.

    Along this thread of research, researchers attempt to incorporate local-patterns into the Transformer model. These works could be broadly categorized into two aspects. One is to enhance the self-attention mechanism via varying the receptive filed across heads, such as convolutional self-attention \cite{yang-etal-2019-convolutional}, multiscale self-attention \cite{guo-2020-msan}, multi-granularity self-attention \cite{hao-etal-2019-multi}. The other is placing the convolution networks in sequence \cite{gulati2020conformer} or parallelly \cite{Zhao-2019-MUSE} in self-attention. The core idea is to endow the Transformer to learn multiscale features rather than the single global interaction.

    Multiscale feature hierarchy is one of the mainstream paradigms to handle the high-resolution tasks in CV \cite{lin2017feature}. More recently, Transformer has been successfully applied to the computer vision area, such as image classification (\textsc{ViT}) \cite{dosovitskiy2020ViT}, object-detection (\textsc{DETR}) \cite{carion2020DETR}.  In particular, multiscale vision Transformers has received great attention due to its superior performance and computation efficiency \cite{wang2021pvt,fan2021MViT,li2021improvedMViT}. The input of ViT is a sequence of patches partitioned from an image, which is similar to the sequential input in NLP systems. But these systems follow a multi-stage paradigm where the channel is increased as the resolution is decreased.

    In this work, we revisit the design of multiscale Transformer based for sequence generation. Although the aforementioned Transformer variants have made great efforts to boost the models’ performance, most of existing works ignore the word-boundaries when modeling the local features. This is because the input sequence is segmented into a series of sub-words \cite{sennrich-subword-neural,kudo-richardson-2018-sentencepiece} to prevent these systems from the out-of-vocabulary problem. However, most of previous techniques, e.g. convolutional self-attention \cite{yang-etal-2019-convolutional}, multiscale self-attention \cite{guo-2020-msan}  directly enhance the feature extraction of neighboring elements but simply discard those language units that naturally exist. This leads to redundant and ambiguous attention map as the relationship between sub-words and words is lack of interpretability.

    Our main contributions could be summarized as follows:
    \begin{itemize}
        \item We re-define the concept of scale for NLP, including scales of sub-word, word and phrase. Our intention is to leverage the word boundaries and phrase-level prior knowledge to compensate for the sub-word features.
        \item Given a sequence, each token is regarded as an individual. The correlations among the three scales are described in three-fold: inter-individual, intra-group and inter-group. We show that learning multiscale features is helpful.
        \item We propose an \textbf{U}niversal \textbf{M}ulti\textbf{S}cale \textbf{T}ransformer, namely \textsc{Umst} - a universal approach to extracting features from different scales. Also, \textsc{Umst} is flexible and  provides an opportunity to incorporate other prior knowledge.
    \end{itemize}
    We comprehensively evaluate the proposed \textsc{Umst} model on the commonly used sequence generation tasks, including machine translation and abstractive summarization. Experimental results show that \textsc{Umst} achieve favorable results than the Transformer and recently multiscale Transformer variants without sacrificing the efficiency. Notably, it yields 0.88 and 0.44 BLEU point gains compared with Transformer under the base and big configurations. It significantly outperforms the Transformer over 1 Rouge point gain in average. \textsc{Umst} is orthogonal to the low-level modeling techniques, e.g. it gains an additional improvement of 0.4 BLEU points when working with relative positional representations \cite{shaw-etal-2018-self}.

    \section{Preliminary}
    \label{sec:background}
    Transformer \cite{vaswani2017attention} follows the encoder-decoder paradigm \cite{sutskever2014sequence}. There are six identical blocks with individual initialization on both the encoder and decoder sides. A standard encoder block consists of two operations. The first is a multi-head self-attention operation for modeling the correlations over all positions. The other is a two-layer feed-forward network for modeling relations within an element in a point-wise fashion. Intertwining these operations with the residual connection \cite{he2016deep} and layer normalization \cite{lei2016layer} allows transformers to be optimized stably.

    Here, we take the Pre-Norm Transformer \cite{wang-etal-2019-learning} as an instance. Let $x$ and $y$ be the input and output. A Pre-Norm Transformer block can be described as: $y = \bar{x} + \mathrm{FFN}(\mathrm{LN(\bar{x})})$ and $\bar{x} = x + \mathrm{SAN}(\mathrm{LN(x)})$.
    $\mathrm{SAN}$ and $\mathrm{FFN}$ denote the multi-head attention sub-layer and the feed-forward sub-layer, respectively. $\mathrm{LN}$ denotes the layer normalization function, where $\textrm{LN}(x) = \frac{x-\mu}{\delta} \circ \gamma + \beta$. $\mu$ and $\delta$ are the mean and standard deviation of a feature, and $\circ$ denotes the element-wise dot.
    \input{./Figure/matrix_final.tex}

    Given the input sequence $x \in \mathbb{R}^{L \times d}$, where $L$ denotes the sequence length and $d$ denotes the hidden size. SAN first obtains the transformation matrix $Q$, $K$, $V$ via three sets of projections, where $Q=xW_{q}$, $K=xW_{k}$ and $V=xW_{v}$. $W_q$, $W_k$ and $W_v$ are trainable parameters. Then, SAN models the global correlations among all positions via $\mathrm{SAN} = \mathrm{ATTN} \cdot V$, where $\mathrm{ATTN}$ can be obtained by:
    \begin{equation}
      \label{eq:attn_map}
      \mathrm{ATTN} = \mathrm{Softmax}(\frac{Q \cdot K^{\mathsf{T}}}{\sqrt{d_k}})
    \end{equation}
    \noindent For simplification, we take the number of heads to 1 as an instance. Then FFN is formalized as: $\mathrm{FFN} = \textrm{\texttt{max}}(\bar{x}W_{1} + b_1, 0)W_2 + b_2$, where $W_1$ and $W_2$ are transformation matrices and $b_1$ and $b_2$ are bias.

    \input{./Figure/architecture_remove_ln_resize.tex}
    \subsection{Definition of Scales and Relations}
    \label{sec:matrix}
    We start with the definition of linguistic-partitioned scale in this work. In natural language processing, it is natural to see a text in multiple levels, e.g. sub-words, words, phrases, sentences, paragraphs. We adopt this idea to define our linguistic-partitioned scales. In this work, we take sub-words, words and phrases as the basis of these scales. The sub-words are the lowest-level scale while the phrases are the highest-level scale.

     We then define the relations among different scales. Take the byte-pair encoding (BPE) \cite{sennrich-subword-neural} as an instance. A BPE-partitioned sequence composed of multilevel contents, e.g. sub-words, words, and phrases. We regard a sub-word as an individual, and words as a group which composed of the corresponding sub-words. Note that some words share both identities mentioned above. Thus, we have the following relations:
    \begin{itemize}
        \item \textbf{Inter-individual relation.} It describes the relationship among input tokens, consisting of sub-words and words. This is the global relation in standard Transformer.
        \item \textbf{Intra-group relation.} It is a local fine-grained relation which endows sub-words to obtain the corresponding word boundary information.
        \item \textbf{Inter-group relation.} It is obtained by the dependency parsing which describes the relationship among groups.
    \end{itemize}

    Our intention is to use the intra-group and inter-group relations to compensate for the inter-individual interactions. To facilitate the information flow between them, we define three transformation matrices, $\mathcal{A}_{w}$, $\mathcal{A}_{p}$ and $\mathcal{G}_{b\rightarrow w}$, where $\mathcal{A}_{w}$ and $\mathcal{A}_{p}$ denote the adjacency matrices of the individual and group, respectively. $\mathcal{G}_{b\rightarrow w}$ is a transformation matrix that indicates the mapping between the word (group) and its corresponding sub-words (individual). Figure \ref{fig:matrix} gives an example. Given a sequence $x \in \mathbb{R}^{L\times d}$, we define three matrices for describing the scales.

    $\mathcal{A}_{w}$ is the adjacency matrix in $\mathbb{R}^{L \times {L}}$ which satisfies:  $\mathcal{A}^{i,j}_{w} = 1$ if and only if the sub-tokens $i$ and $j$ belong to the same word. $\mathcal{A}_{w}$ is a block diagonal matrix where entries in the same color denote a group of sub-tokens of a word.

    $\mathcal{A}_{p}$ is an adjacency matrix in $\mathbb{R}^{L^{'} \times {L^{'}}}$. It satisfies:  $\mathcal{A}^{i,j}_{p} = 1$ if only if the tokens $i$ and $j$ have a dependency edge. As is seen in Figure \ref{fig:word2wordmatrix}, dashed lines denote the dependency between two words, which can be easily obtained through open-source parsing tools, e.g. StanfordNLP \cite{qi2018universal}, Berkeley Neural Parser \cite{kitaev-klein-2018-constituency}. Notably, we build $\mathcal{A}_{p}$ as an undirected graph. It is a symmetry matrix where corresponding elements are shaded with different colors. Similar with $\mathcal{A}_{w}$, it portrays the relation among words rather than sub-words. 

    $\mathcal{G}_{b \rightarrow w}$ is a matrix in $\mathbb{R}^{L \times {L}^{'}}$.  Its function is to solve the spatial size mismatch problem between word level and sub-word level. $\mathcal{G}_{b\rightarrow w}$ satisfies:  $\mathcal{G}^{i,j}_{b\rightarrow w} = 1$ if only if the sub-token $i$ is belonging to token $j$. $\mathcal{G}_{b\rightarrow w}$ is an orthogonal matrix of column vectors which guarantees the lossless rank of the matrix.

    \section{Universal Multiscale Transformer}
    \label{sec:UMST}
    As noted in \Cref{sec:background}, SAN collects the interaction information over all positions, no matter if a token is a word or a sub-word. This results in ineffective interactions and ambiguous attention map. To remedy this issue, we redesign the Transformer architecture from the multiscale perspective. The overall architecture is depicted in Figure \ref{fig:architecture}. As we can see, it slightly differs from the standard Transformer (Figure \ref{fig:architecture} (left)) since we employ two additional graph convolutional networks before computing the attention. The core of this design is to coverage the intra-group and inter-group interactions via W-GCN and P-GCN, respectively. Also, we replace the standard self-attention network by the proposed rectified self-attention mechanism.

    \subsection{Class Embedding}
    \label{sec:class_embedding}
    The standard Transformer receives a sequence of embedding as input. The embedding is a mixture of token embedding and position embedding. In order to distinguish which part of position implies sub-tokens, we introduce a binary class embedding, where BPE-level tokens have a value of 1, and word-level tokens have a value of 0. The embedding is randomly initialized by a normal distribution where the mean is 0 and the standard deviation is $1/\sqrt{d}$.

    \begin{table}[t]
    \centering
    \caption{Several operations to handle the local-pattern.}
    \label{tab:operation}
    \setlength{\tabcolsep}{2.0pt}
    \vskip 0.15in
    \small
    \begin{tabular}{lccc}
    \toprule
    Operations      & Efficiency & Expression & Stable \\
    \midrule
    Average Pooling & \checkmark &    & \checkmark  \\
    GAT \cite{Petar2018GAT} &       &   \checkmark    &        \\
    GCN \cite{Kipf2017gcn} & \checkmark      &   \checkmark    &   \checkmark     \\
    \bottomrule
    \end{tabular}
    \end{table}

    \subsection{Intra-Group Interaction}
    \label{sec:intra_group}
    Previous studies ignore the word-boundary information and directly modeling the local feature in a fixed window, which results in degeneration of expression and low interpretability. To tackle these problems, we firstly model the intra-group correlations where sub-words belonging to the same word have a high similarity among them.

    Given the word boundary $\mathcal{A}_{w}$, there are several ways to gain the fusion representation, such as average pooling, graph attention networks (GAT) \cite{Petar2018GAT}, and graph convolutional networks (GCN) \cite{Kipf2017gcn}. Table \ref{tab:operation} summarizes the properties of three operations considering the expression power, training stability and computation efficiency. In our preliminary experiments, the average pooling method is light and stable while having a poor expression ability. On the contrary, graph attention networks show considerable power to capture the intra-group dependency. However, stacking several attention-based modules leads to poor stability during training, especially when our model becomes deep. Finally, we choose GCN as the default model by taking these factors into account.

    A GCN layer encapsulates each node's hidden representation by aggregating feature information from its neighbors. Our intention is to let each sub-words receive its corresponding word-level information.
    We assume a sequence $x \in \mathbb{R}^{L\times d}$. Instead of feeding the input into SAN, we use a GCN \cite{Kipf2017gcn} to enhance the feature extraction within each group. The formulation of GCN can be described as:
    \begin{equation} \label{eq:gcn-word}
      \mathrm{GCN}_{\mathrm{word}}  = \sigma(\tilde{D}_w^{-\frac{1}{2}} \tilde{\mathcal{A}_{w}} \tilde{D}_w^{-\frac{1}{2}} \cdot x W_w)
    \end{equation}
    \noindent $\tilde{\mathcal{A}_{w}}={\mathcal{A}_{w}}+\mathcal{I}_{L}$ the adjacency matrix of the undirected graph with self-connections. Here $\mathcal{I}_L$ denotes the identity matrix. $\tilde{D}_w$ is the degree matrix of the adjacency matrix $\tilde{\mathcal{A}_w}$. $W_w$ is a linear transformation which is a trainable parameter. $\sigma$ denotes an activation function. We choose ReLU for the design of activation function: $\textrm{ReLU}=\mathrm{max}(0,\cdot)$. To further strengthen the fine-grained feature, we multiply the output of $\mathrm{GCN}_{\mathrm{word}}$ by $\tilde{D}_w^{-\frac{1}{2}} \tilde{\mathcal{A}_{w}} \tilde{D}_w^{-\frac{1}{2}}$ at the left side.

    Essentially, the output $\mathrm{GCN}_{\mathrm{word}}$ still has a length $L$ which is not changed. In this way, each sub-words not only keeps the original meaning, but also derives the corresponding word boundaries.

    \input{./Figure/attention_map_trans.tex}
    \subsection{Inter-Group Interaction}
    \label{sec:inter_group}
    Beyond the intra-group interaction, another conceive here is to make full use of the phrase-level prior knowledge. The dependency among words portrays a long-range relationship rather a local constraint within a varying window size. It is a kind of phrase-level prior which could be easily obtained through open-source tools. Notably, the phrase prior is built upon words rather than sub-words. Thus we need to down-sample the representation $X_{word}^{L^{'} \times d}$ from the length $L$ into $L^{'}$. Here, we choose the averaging pooling operation to aggregate into the word-level embedding. Subsequently, we model the inter-group interaction via the adjacency matrix $\mathcal{A}_{p}$, where the formation is as follows:
    \begin{equation} \label{eq:gcn-phrase}
      \mathrm{GCN}_{\mathrm{phrase}} = \sigma(\tilde{D}_p^{-\frac{1}{2}} \tilde{\mathcal{A}_{p}} \tilde{D}_p^{-\frac{1}{2}} \cdot x W_p)
    \end{equation}
    \noindent where $\tilde{D}_p$ is the degree matrix of the adjacency matrix $\tilde{\mathcal{A}_{p}}$. $W_p$ is a learnable linear projection. One can share the projection matrices $W_w$ and $W_p$ across all transformer blocks, which reduces the total parameters in a certain extent. Another choice is to use a shared projection matrix when handling the word-level GCN and the phrase-level GCN, but it slightly hurts the performance. All these discussions could be found in \cref{sec:analysis}.

    \subsection{Rectified Self-attention}
    \label{sec:rsan}
    The proposed \textsc{Rsan} module is a rectified multi-head self-attention via a double-branch manner. Formally, there are two kinds of input. The first input $X_{bpe}^{L \times d}$ contains the word-level interaction, and the other input $X_{word}^{L^{'} \times d}$ is further enhanced by the phrase-level interaction. As depicted in Figure \ref{fig:architecture}, the right branch (yellow) is similar with the original self-attention which computes the correlations over all positions. The attention map $\mathrm{ATTN}_1 \in \mathbb{R}^{L \times L}$ could be easily obtained through Eq. \ref{eq:attn_map}, where the query and key are generated from $X_{bpe}^{L \times d}$. The left branch (blue) aims to extract higher-level inter-group interaction over all words. Likewise, we can obtain the attention map $\mathrm{ATTN}_\mathrm{2} \in \mathbb{R}^{L^{'} \times L^{'}}$.

    After obtaining two distributions, a next step is converting the attention map from $L^{'} \times L^{'}$ into $L \times L$ using the following transformation:
    \begin{equation}
        \mathrm{ATTN}_2^{'} = \mathcal{G}_{b\rightarrow w}\cdot \mathrm{ATTN}_2 \cdot (\mathcal{G}_{b\rightarrow w})^{\mathsf{T}}
    \end{equation}
    \noindent where $\mathcal{G}_{b \rightarrow w}$ is the transformation mapping. The data follow is shown in Figure \ref{fig:matrix_trans}. The attention map $\mathrm{ATTN}_1$ portrays an inter-individual correlation whose $Q_{bpe}^{L \times d}$ and $K_{bpe}^{L \times d}$ have already been augmented via the injected word-boundaries. Meanwhile, the attention map $\mathrm{ATTN}_2$ captures a high-level interaction among each word rather than the mixture of sub-words and words. Our intention here is to use $\mathrm{ATTN}_2$ to enhance $\mathrm{ATTN}_1$.

    There are several approaches to attaining the final attention map, e.g. sum of the two distributions, average pooling or gated mechanism. However, there is no significant performance gap in our preliminary experiments. Here, we choose the average pooling to guarantee the normalization.

    The final output of the proposed rectified self-attention is the multiplication of the fused attention map $\mathrm{ATTN}_2^{'}$ with value $V_{bpe}^{L \times d}$. In this way, \textsc{Rsan} incorporate both the word-boundaries and the phrase-level prior knowledge.
    \section{Experimental Setups}
    \label{sec:exp}
    We evaluate the proposed multiscale Transformer on two sequence generation benchmarks, including machine translation and abstractive summarization.

    \subsection{Dataset}
    \paragraph{Machine Translation}
     We report results on two machine translation datasets, including a large-scale WMT'14 English-German (En-De) dataset, and a WMT'16 English-Romanian (En-Ro) dataset. For the En-De dataset, the training data consisted of approximately $4.5$M tokenized sentence pairs, as in \cite{vaswani2017attention}. We preprocess the dataset following the same setup in \citet{ott2018scaling}'s work, obtaining a clean and high-quality bilingual training data containing $3,894,376$ sentences. We selected \textit{newstest2013} as the validation data and \textit{newstest2014} as the test data. To validate the proposed method on various granularities, we segmented the sentences into sub-word units \cite{sennrich-subword-neural} with $7$K \cite{xu-etal-2021-vocabulary}, $15$K and $32$K merging operations. For the En-Ro dataset, it consisted of $610$K bilingual sentence pairs. Also, we adopted the same scripts as those used in  \citet{lee-etal-2018-deterministic,kasai2020disco}'s work. We adopt a joint source and target BPE factorization with the vocabulary size of $40$K. We use newsdev-2016 and newstest-2016 as the validation and test sets, respectively.

    \begin{table}[t]
        \caption{Comparison with previous studies on the WMT En-De task. The results with $\dag$ denote our re-implementing results, since some of works have not been evaluated on MT.}
        \label{tab:main-results}
          \vskip 0.15in
          \small
          \setlength{\tabcolsep}{1pt}
          \centering
          \begin{tabular}{lrrrr}
          \midrule
          \multirow{2}{*}{\textbf{Model}}  & \multicolumn{2}{c}{\textbf{Base}} & \multicolumn{2}{c}{\textbf{Big}}  \\
          \cmidrule(r){2-3} \cmidrule(r){4-5}
          &  \bf Param & \bf BLEU & \bf Param & \bf BLEU \\
            \midrule
            Transformer     \cite{vaswani2017attention}   &65M   &27.30   &213M  &28.40 \\
            Scaling NMT     \cite{ott2018scaling}         &-     &-       &210M  &29.30 \\
            \textsc{Dlcl}   \cite{wang-etal-2019-learning}&62M   &27.30   &-     &- \\
            \textsc{Muse}   \cite{Zhao-2019-MUSE}         &-     &-       &-     &29.90 \\
            \textsc{Mg-Sa}  \cite{hao-etal-2019-multi}    &89M   &28.28   &272M  &29.01 \\
            \midrule
            Transformer $\dag$                            &65M   &27.63   &216M   &29.31  \\
            \textsc{Muse}$\dag$ \cite{Zhao-2019-MUSE}     &68M   &27.97   &233M   &29.11  \\
            \textsc{MSMSA}$\dag$ \cite{guo-2020-msan}     &65M   &27.57   &233M   &28.84  \\
            \textsc{TNT}$\dag$ \cite{han2021transformer}  &83M   &28.48   & -     & -     \\
            \textsc{Umst}                                 &70M   &28.51   &242M   &29.75  \\
            \textsc{Umst} + RPR                           &70M   &28.90   &242M   &30.15  \\
            \bottomrule
        \end{tabular}
    \end{table}

    \paragraph{Abstractive Summarization}
    We also test the models's ability to process long sequences on the CNN-DailyMail summarization task \cite{nallapati-etal-2016-abstractive,hermann2015teaching}. The preprocessed method is the same as in \cite{ott2019fairseq}'s work. We use a shared BPE with $10$K operations, resulting in a vocabulary of $10,420$ entries.

    \subsection{Implementation Details}
    As discussed in \Cref{sec:inter_group}, here we choose the open-source parsing tool proposed by Stanford \footnote{\url{https://stanfordnlp.github.io/CoreNLP/}} to extract the dependency tree. Due to the page limit, more experimental setups and model configurations could be found in \Cref{sec:exp_setup}.

    \section{Results}
    \label{sec:results}
    \subsection{Machine Translation}
    \label{sec:mt}
    \paragraph{Results on WMT En-De}
    Table \ref{tab:main-results} compares the proposed \textsc{Umst} with several strongly related systems. The results are evaluated on WMT'14 En-De newstest2014 test set. We re-implement these methods in our codebase within the same setting for fair comparisons.
    First, we can see that the standard Transformer delivers a BLEU point of 27.63 and 29.31 under the base/big configurations. Our \textsc{Umst} outperforms it by 0.88 and 0.44 BLEU points, indicating that incorporating word-boundaries and phrase-level prior knowledge is indeed helpful. Also, \textsc{Umst} is orthogonal to previous local modeling methods \cite{yang-etal-2019-convolutional, shaw-etal-2018-self,Zhao-2019-MUSE}. Here, we take RPR \cite{shaw-etal-2018-self} as an instance. When applying RPR into \textsc{Rsan}, our model can obtain an additional improvement of 0.4 BLEU points. Notably, \textsc{Umst} can beat previous multiscale variants under the same experimental setting. For example, it can beat TNT \cite{han2021transformer} whose parameters equals to a 12-layer Transformer. This indicates that redefining scales from the linguistic unit perspective is valuable. More details and comparisons could be found in \cref{sec:appendix_exp}.
    \begin{table}[t]
      \caption{Results on the WMT En-Ro task.}
      \label{tab:en-ro}
      \vskip 0.15in
      \small
      \setlength{\tabcolsep}{5.0pt}
      \centering
      \begin{tabular}{lrr}
      \toprule
      \textbf{Model} & \bf Param & \bf BLEU \\
      \midrule
      \textsc{Delight} \cite{mehta2020delight}     & 53M & 34.70  \\
      Baseline in \textsc{mBART} \cite{liu-etal-2020-multilingual-denoising}          & -   & 34.30 \\
      Baseline in \textsc{Disco} \cite{kasai2020disco}          & -   & 34.16 \\
      Transformer $\dag$ \cite{vaswani2017attention}             & 54M  & 34.21      \\
      \textsc{TNT}$ \dag$  \cite{han2021transformer}              & 73M  &34.00       \\
      \textsc{Umst}                                    & 60M  & \bf 34.81   \\
      \textsc{Umst} + RPR                              & 60M  & \bf 35.31   \\
      \bottomrule
    \end{tabular}
    \end{table}
    \begin{table}[t]
      \caption{Results of the proposed \textsc{Umst} and previous studies on the CNN-DailyMail dataset.}
      \label{tab:summarization}
      \vskip 0.15in
      \small
      \setlength{\tabcolsep}{2.7pt}
      \centering
      \begin{tabular}{lccc}
      \toprule
      \textbf{Model} & \bf RG-1 & \bf RG-2 & \bf RG-L \\
      \midrule
      \textsc{DynamicConv} \cite{wu2019DynamicConv}    & 39.84  & 16.25  & 36.73  \\
      \textsc{Bottom-Up} \cite{gehrmann-etal-2018-bottom} & 41.22 & 18.68 &38.34 \\
      \textsc{Surface} \cite{liu2020understanding}     & 41.00  & 18.30    & 37.90  \\
      \textsc{Dman}  \cite{fan-etal-2021-mask}         & 40.98  & 18.29    & 37.88  \\
      \midrule
      Transformer$\dag$                 & 40.55  & 17.81    & 37.47  \\
      \textsc{Umst} w/o inter-group interactions   & 41.62  & 18.65    & 38.28  \\
      \textsc{Umst}                                    & \bf 41.82  & \bf 18.91    & \bf 38.54  \\
      \bottomrule
    \end{tabular}
    \end{table}
    \paragraph{Results on WMT En-Ro}
    Similarly, \textsc{Umst} yileds a 34.81 BLEU point on the WMT En-Ro task (see Table \ref{tab:en-ro}), outperforming the previous results in terms of BLEU. Also, the RPR-enhanced \textsc{Umst} almost achieves the state-of-art with no use of pre-training models, it outperforms \textsc{Umst} by 0.5 BLEU points. We will further validate the combination with other local modeling methods in future work.

    \subsection{Abstractive Summarization}
    The results on abstractive summarization are listed in Table \ref{tab:summarization}. Note that we choose the base configuration, and all systems consist of 6 blocks for both encoder and decoder. We can see that the proposed \textsc{Umst} outperforms the standard Transformer by a large margin (e.g. 1.27 Rouge-1 benefits) as the the decoupled multiscale attention receives benefits from both the intra-group (word boundaries) and inter-group (phrase prior) interactions. Note that the model can still attain nearly 1 rouge gains in terms of three metrics when removing the phrase-level prior knowledge. This observation again demonstrates the essential of word-boundaries.

    \begin{table}[t]
      \caption{The ablation study on the WMT En-De testset.}
      \label{tab:ablation_study}
      \vskip 0.15in
      \small
      \setlength{\tabcolsep}{2.5pt}
      \centering
      \begin{tabular}{lccccc}
      \toprule
      \textbf{Model} &  \bf Depth &  \bf BLEU &  \bf Depth &  \bf BLEU \\
      \midrule
      Transformer                   & 6-6 & 27.63 & 12-6      & 28.67  \\
      \textsc{Umst}                 & 6-6 & 28.51 & 12-6      & 29.49  \\
      w/o class-embedding           & 6-6 & 28.39 & 12-6      & 28.99  \\
      w/o intra-group interactions  & 6-6 & 27.87 & 12-6      & failed \\
      w/o inter-group interactions  & 6-6 & 28.06 & 12-6      & 29.37  \\
      replace GCN with pooling      & 6-6 & 27.96 & 12-6      & 28.89 \\
      replace GCN with GAT          & 6-6 & 28.11 & 12-6      & failed \\
      \bottomrule
    \end{tabular}
    \end{table}
    \begin{table}[t]
      \caption{Comparisons of shared parameters and several mappings on the WMT En-De task.}
      \label{tab:para_sharing}
      \vskip 0.15in
      \small
      \setlength{\tabcolsep}{3.0pt}
      \centering
      \begin{tabular}{llrr}
      \toprule
      \textbf{\#} &\textbf{Model} & \bf Param & \bf BLEU \\
      \midrule
      0  &Transformer                              & 65M   & 27.63         \\
      1  &\textsc{Umst}                            & 70M   & \bf 28.51     \\
      2  &shared $W_w$ and $W_p$ (across blocks)   & 66M   & 28.33         \\
      3  &shared $W_w$ and $W_p$ (within each block)& 69M   & 28.39         \\
      4  &shared $Q$, $K$ in \textsc{Rsan}                  & 67M   & 28.41         \\
      \midrule
      5  &replace by nearest interpolation      & 70M   & 27.96         \\
      6  &replace by linear interpolation       & 70M   & 27.61         \\
      \bottomrule
    \end{tabular}
    \end{table}

    \section{Analysis}
    \label{sec:analysis}
     In this section, we present a detailed analysis to provide some insights on why \textsc{Umst} improves over Transformer.
    \definecolor{tiffanyblue}{RGB}{129,216,208}
\definecolor{bangdiblue}{RGB}{0,149,182}
\definecolor{kleinblue}{RGB}{0,47,167}
\definecolor{kabuliblue}{RGB}{26,85,153}
\definecolor{purple}{RGB}{138,43,226}

\begin{figure*}[!t]
  \centering
  \begin{tikzpicture}[]
    \scriptsize{
      \begin{axis}[
	 at={(0,0)},
      ymajorgrids,
      xmajorgrids,
      grid style=dashed,
      width=.30\textwidth,
      height=.24\textwidth,
      legend style={at={(0.38,0.13)}, anchor=south west},
      xlabel={\scriptsize{Encoder Depth}},
      ylabel={\scriptsize{BLEU}},
      ylabel style={yshift=-2em},xlabel style={yshift=0.0em},
      yticklabel style={/pgf/number format/precision=1,/pgf/number format/fixed zerofill},
      ymin=27,ymax=30.5, ytick={27,28,29,30},
      xmin=4,xmax=26,xtick={6,12,18,24},
      legend style={yshift=-6pt,xshift=-1em, legend plot pos=right,font={\footnotesize},cells={anchor=west}}
      ]

        
      \addplot[red,mark=pentagon*,,mark size=2.5pt,thick,mark options={fill=white,draw=red,line width=1.0pt}] coordinates {(6,27.63) (12,28.67) (18,29.30) (24,29.5)};
      \addlegendentry{\scalebox{.9}{{Transformer}}}


      \addplot[kleinblue,mark=diamond*,mark size=2.5pt,thick,mark options={fill=white,draw=kleinblue,line width=1.0pt}] coordinates {(6,28.51) (12,29.49) (18,29.81) (24,30.15)
      };
      \addlegendentry{\scalebox{.9}{\textsc{Umst}}}



      \end{axis}
     }

	\scriptsize{
      \begin{axis}[
	 at={(20em,0)},
      ymajorgrids,
      xmajorgrids,
      grid style=dashed,
      width=.30\textwidth,
      height=.24\textwidth,
      legend style={at={(0.38,0.13)}, anchor=south west},
      xlabel={\scriptsize{Number of BPE (K)}},
      ylabel={\scriptsize{BLEU}},
      ylabel style={yshift=-2em},xlabel style={yshift=0.0em},
      yticklabel style={/pgf/number format/precision=1,/pgf/number format/fixed zerofill},
      ymin=26.7,ymax=28.8, ytick={27.00, 27.50, 28.00,28.50},
      xmin=1,xmax=35,xtick={0,10,20,30,40},
      legend style={yshift=-6pt,xshift=-1em, legend plot pos=right,font={\footnotesize},cells={anchor=west}}
      ]
      \addplot[red,mark=pentagon*,mark size=2.5pt,thick,mark options={fill=white,draw=red,line width=1.0pt}] coordinates { (7,27.63)  (15,27.82) (32,27.84)
      };
      \addlegendentry{\scalebox{.9}{{Transformer}}}

      \addplot[kleinblue,mark=diamond*,mark size=2.5pt,thick,mark options={fill=white,draw=kleinblue,line width=1.0pt}] coordinates {(7, 28.51) (15,28.56) (32,28.5)};
      \addlegendentry{\scalebox{.9}{\textsc{Umst}}}
      \end{axis}

     }
     \scriptsize{
     \begin{axis}[
     at={(40em,0)},
      width=.30\textwidth,
      height=.24\textwidth,
      legend style={at={(0.43,0.47)}, anchor=south west},
      xlabel={\scriptsize{Number of BPE}},
      ylabel={\scriptsize{Number-of-Stentences(M)}},
      ylabel style={yshift=-2em},xlabel style={yshift=0.0em},
      yticklabel style={/pgf/number format/precision=1,/pgf/number format/fixed zerofill},
      xmin=0, xmax=25,
      ymin=0, ymax=1200000,
      xtick={0,5,10,15,20,25},
      ytick={0,200000,400000,600000,800000,1000000,1200000},
      yticklabels={0,0.2,0.4,0.6,0.8,1.0,1.2},
      ymajorgrids,
      xmajorgrids,
      grid style=dashed,
      legend cell align=left,
      scaled ticks=false,
      y tick style={opacity=0},
    legend style={yshift=-1.3pt,xshift=-0.9em, legend plot pos=right,font={\footnotesize},cells={anchor=west}}
      ]
      \addplot [sharp plot,darkgreen,thick,mark size=1.5pt ,line width=0.5pt,mark size=0.5pt] table [x=Number-of-BPE,y=Number-of-Stentences,col sep=comma] {Figure/Data/1-1-best-paper-data.csv};
      \addlegendentry{\scalebox{.8}{BPE-7K}}
      \addplot [sharp plot,goldenorange,mark size=1.5pt,thick,line width=0.5pt,mark size=0.2pt] table [x=Number-of-BPE,y=Number-of-Stentences,col sep=comma] {Figure/Data/volt-15k-data.csv};
      \addlegendentry{\scalebox{.8}{BPE-15K}}
      \addplot [sharp plot,royalblue,thick,mark size=1.5pt ,line width=0.5pt,mark size=0.2pt] table [x=Number-of-BPE,y=Number-of-Stentences,col sep=comma] {Figure/Data/volt-32k-data.csv};
      \addlegendentry{\scalebox{.8}{BPE-32K}}

  \end{axis}
     
     }
  \end{tikzpicture}
  \caption{The comparison of BLEU against different encoder depths and BPE merging operations.}\label{fig:encoder-detph}
\end{figure*}
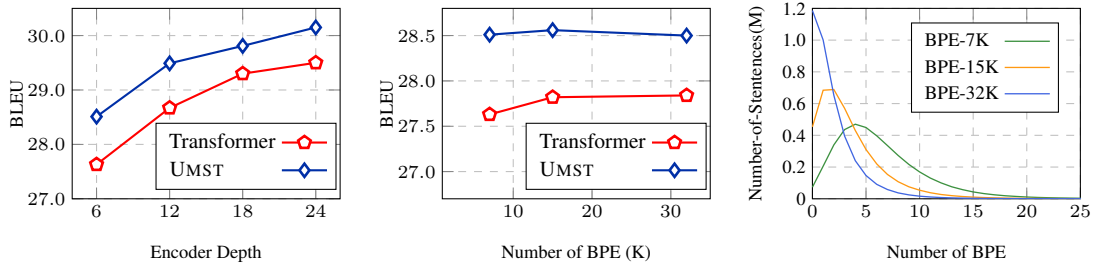
    \paragraph{Ablation studies}
    We conduct a series of ablation studies to inspect \textsc{Umst} from different aspects. Table \ref{tab:ablation_study} summarizes the results of depth 6-6 and 12-6 on the WMT'14 En-De task. First, as discussed in  \Cref{sec:class_embedding}, the proposed class embedding is used to distinguish whether the current token is a word or a sub-word. Through the results, we find it plays an quite important role when the model goes deep. There is a 0.5 BLEU point drop when removing it.

    In this work, both intra-group and inter-group interaction could be regarded as two kinds of local-pattern modeling. The former is to incorporate the word-boundaries, which acts within a continuous restricted window (inside a word). While the latter is a sparse and long-range dependency, containing high-level semantic information. Through the results we find that removing either of them causes BLEU degradations, especially for the intra-group interaction. It results in unstable training when optimizing a 12-layer system. This is because that inter-group interaction is established upon words, rather than sub-words. So UMST w/o intra-group interactions ignoring the word-boundary information brings noise signals to the training and even destroys the training stability. Moreover, inter-group interaction is more important for the shallow model. It causes a 0.43 BLEU drop when removing the inter-group interaction. However, only a 0.12 BLEU drop could be observed for deeper models. This is because deep models may gradually gain high-level semantic features through the stacked self-attention.
    
    \begin{table}[t]
      \caption{Discussion on training and inference efficiency. The experiments are conducted based on \texttt{Transformer-Base}.}
      \label{tab:efficiency}
      \vskip 0.15in
      \setlength{\tabcolsep}{3.5pt}
      \small
      \centering
      \begin{tabular}{lccc}
        \toprule
        \bf Model  & \bf Inference & \bf Training  & \bf GPU \\
        \midrule
        Transformer                & 142.33    & 1.00$\times$  &  6.4G\\
        \textsc{Umst} w/o word     & 147.60    & 1.18$\times$  & 6.5G\\
        \textsc{Umst}              & 136.67    & 1.25$\times$ & 6.7G\\
        \bottomrule
        \end{tabular}
    \end{table}
    
    We then replace GCN with the average pooling and GAT. We see that replacing GCN with both alternatives suffer from a sharp decrease in performance. The situation becomes more serious for GAT when switching to a deeper scenario, that it even fails to converge. One reasonable explanation is that stacking multiple continuous attention networks raises the optimization difficulty and makes the training be fragile. 

    \paragraph{Effect of sharing parameters}
    As discussed in \Cref{sec:inter_group} and \Cref{sec:rsan}, one can share the projection matrices $W_w$ and $W_p$ across all transformer blocks (\#2, totally 2 matrices), share $W_w$ and $W_p$ within the block (\#3, totally 6 matrices) or use the same series of $Q$ and $K$ transformation matrices inside of \textsc{Rsan} (\#4). Table \ref{tab:para_sharing} shows that these strategies achieve almost comparable results with the default setting but consuming less parameters. And they all outperform the baseline by a large margin.

    \paragraph{Comparisons of up-sampling methods}
    Except for the up-sampling operation depicted in Figure \ref{fig:matrix_trans}, which converts $\mathrm{ATTN}_2$ to $\mathrm{ATTN}_2^{'}$, we also test two well-popular up-sampling methods in CV, including nearest interpolation and linear interpolation. Unfortunately, both methods behave much inferior to ours. Notably, the system with linear interpolation even underperforms the baseline. This observation indicates that the proposed \textsc{Umst} is a well-defined architecture with specifically designed components.
    
    \input{./Figure/random_matrix.tex}
    \paragraph{BLEU against encoder depths}
    Figure \ref{fig:encoder-detph} (left) plots the BLEU scores of \textsc{Umst} and Transformer against different encoder depths. Obviously, we observe \textsc{Umst} beats Transformer under all configurations, attaining almost a 0.76 BLEU gap in average. This observation further demonstrates the effectiveness of the proposed method.

    \paragraph{BLEU against BPE operations}
    Figure \ref{fig:encoder-detph} (middle) compares the performances between \textsc{Umst} and Transformer using different vocabularies. As plotted in Figure \ref{fig:encoder-detph} (right), more sentences are likely to be separated into sub-tokens when a vocabulary gets smaller. In a nutshell, the word-boundary information is more essential within a small vocabulary, where \textsc{Umst} can gain more benefits. Intuitively, \textsc{Umst} achieves consistent BLEU improvements within different-scale BPE merging operations. It gains the largest benefit when the BPE merging operation is 7K.
    
    \begin{figure*}[ht]
	\centering
	\subfigure{
	   \label{fig:attn_map_standard}
	   \begin{minipage}[t]{0.24\linewidth}
        \includegraphics[width=\linewidth]{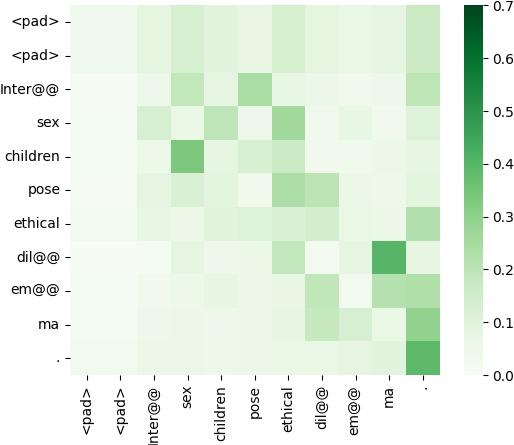}
        \end{minipage}
        \begin{minipage}[t]{0.24\textwidth}
        \includegraphics[width=\linewidth]{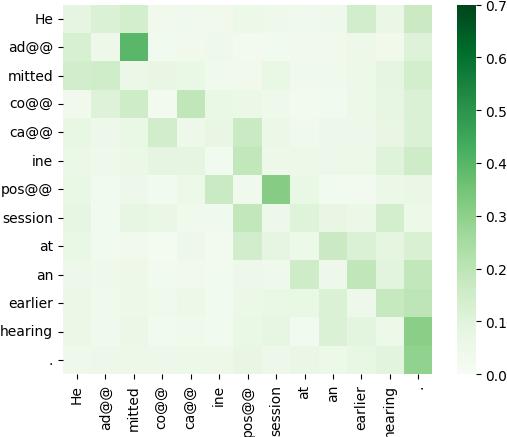}
        \end{minipage}
	    \begin{minipage}[t]{0.24\textwidth}
        \includegraphics[width=\linewidth]{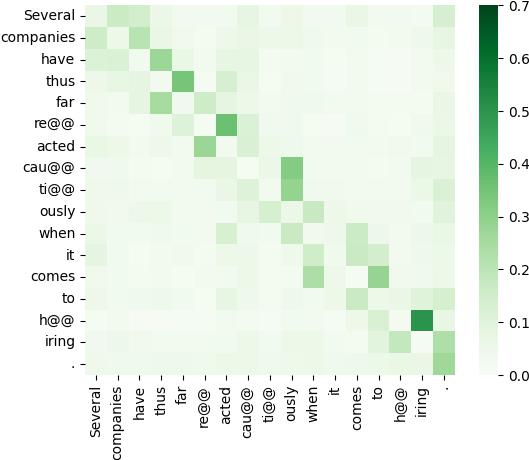}
        \end{minipage}
        \begin{minipage}[t]{0.24\textwidth}
        \includegraphics[width=\linewidth]{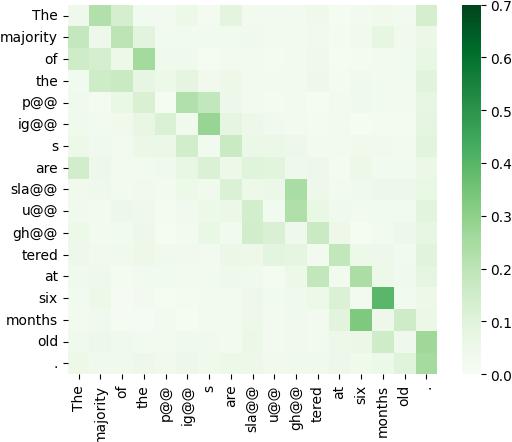}
        \end{minipage}
		
	}%

	\subfigure{
	    \label{fig:attn_map_our}
		\begin{minipage}[t]{0.24\linewidth}
        \includegraphics[width=1\linewidth]{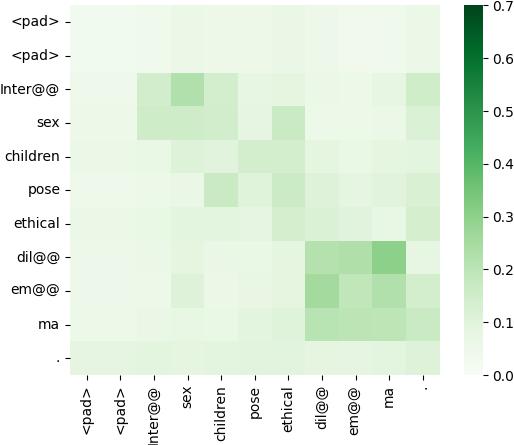}
        \end{minipage}
        \begin{minipage}[t]{0.24\textwidth}
        \includegraphics[width=1\linewidth]{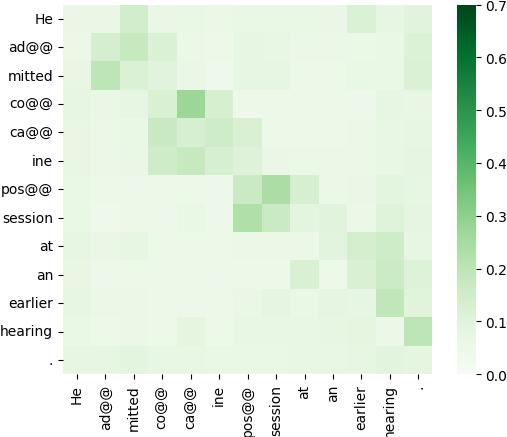}
        \end{minipage}
	    \begin{minipage}[t]{0.24\textwidth}
        \includegraphics[width=1\linewidth]{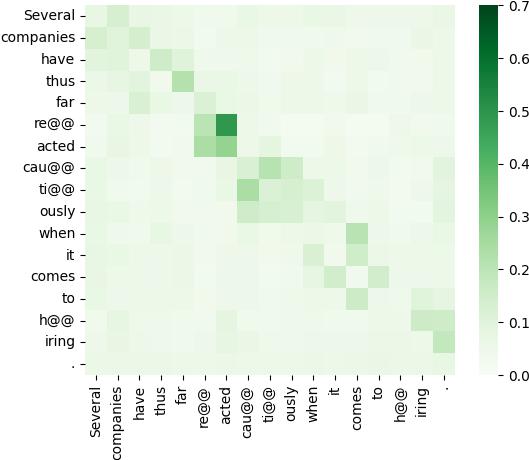}
        \end{minipage}
        \begin{minipage}[t]{0.24\textwidth}
        \includegraphics[width=1\linewidth]{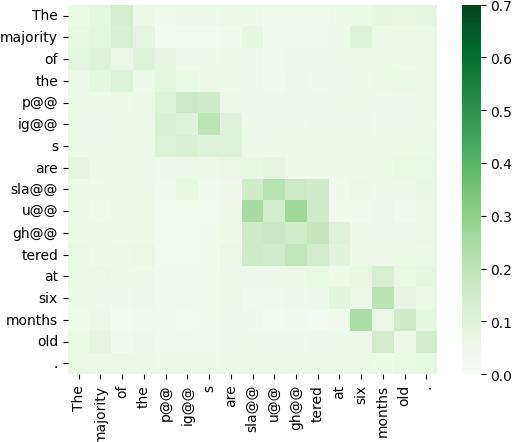}
        \end{minipage}
	}%
	\caption{Quantitative examples of the attention distribution over several real cases. The above is the distribution generated by the standard Transformer, and the below is ours (\textsc{Umst}). Dark color means a higher value in the distribution.}
	\label{fig:attn_map}
\end{figure*}
    
    \paragraph{Discussion on efficiency}
    Deployment of transformer models on devices is important for practical applications. Table \ref{tab:efficiency} summarizes the comparison between \textsc{Umst} and Transformer in terms of the inference speed, training cost and GPU consumption. As we can see that \textsc{Umst} only brings 0.25$\times$ extra training cost, with almost no inference latency, since we only apply our method in the encoder side. This is because the inference overhead mainly comes from the decoder side due to the auto-regressive decoding schema. Also, our \textsc{Umst} only consumes more than 0.3G GPU allocation, which is acceptable.
    
    \paragraph{Evaluation with Other Metrics}
    Previous work has pointed out that BELU may correlate poorly with human judgments. To eliminate the randomization, we also evaluate our proposed \textsc{Umst} on another two evaluation metrics: BLEURT~\cite{sellam-etal-2020-bleurt} and COMET~\cite{rei-etal-2020-comet}. Table \ref{tab:Discussion_on_two_interactions} displays the results. As the comparison of the first two lines in Table \ref{tab:Discussion_on_two_interactions}, we observe the same phenomenon with BLEU that \textsc{Umst} can also outperform the baseline over a large margin on these two metrics. This observation provides a strong evidence for the effectiveness.
    
    \begin{table}
    \vspace{-1.1em}
      \caption{The results in terms of COMET and BLEURT on the WMT En-De task.}
      \vskip 0.15in
      \label{tab:Discussion_on_two_interactions}
      \small
      \setlength{\tabcolsep}{3.0pt}
      \centering
      \begin{tabular}{lccc}
      \toprule
      \textbf{Model}  & \bf BLEU  & \bf COMET & \bf BLEURT\\
      \midrule
      Transformer                           & 27.63/28.67    & 48.24/51.43  &   71.03/72.01 \\
      \textsc{Umst}                         & 28.51/29.49    & 49.04/52.55  &   71.31/72.63\\
      \midrule
      \textsc{Umst} w/o inter-group         & 28.06/29.37    & 48.78/52.42  &   71.21/72.49 \\    
       \quad + Random                & 26.98/27.11        & 46.10/43.04      &    70.54/70.48   \\
       \quad + Shuffle               & 28.08/28.96    & 48.36/51.80     & 71.01/72.38      \\
      \midrule
       \textsc{Umst} + Random P    & 28.20/-    & 48.90/-     & 71.28/-      \\
      \bottomrule
    \end{tabular}
    \end{table}
    
    \paragraph{Random Adjacency Matrices}

    As analyzed above, \textsc{Umst} attains satisfactory performance gains with our well-designed components. However, it is still worthy to figure out whether the gains come from the proper usage of multiscale information or the enlarging model parameters. Here, we make a further in-depth analysis on it to answer this question. A natural idea is to use random adjacency matrices instead of the right one.
    
    We elaborate two manners to construct the random adjacency matrices of $\mathcal{A}_w$ and $\mathcal{A}_p$. For the $\mathcal{A}_w$, we provide two choices, namely  Random $\mathcal{A}_w$ and Shuffle $\mathcal{A}_w$. Random $\mathcal{A}_w$ is a totally random matrix in which each entry is either 1 or 0. However, such a design of $\mathcal{A}_w$ directly breaks the physical logic of BPE partitions, which are composed of several continuous elements. To remedy this problem, we design Shuffle $\mathcal{A}_w$, which is a shuffle of the original BPE partitions. The detailed operations are illustrated in Figure \ref{fig:random_matrix}. 
    
    Table \ref{tab:Discussion_on_two_interactions} summarizes the comparison of two random manners of $\mathcal{A}_w$ \footnote{Note that we drop the inter-group interaction and replace the $\mathcal{A}_w$ with the random versions of $\mathcal{A}_w$}. As we expect, random initialization achieves a much weaker performance, which is even inferior to the baseline. This observations demonstrates the improvement mainly comes from the successful utilization of word-boundary information. On the other hand, shuffling the matrix maintains similar results with pure word-boundary information when the encoder depth is 6. This is because shuffling the matrix enables the model to learn implicit local fine-grained features. As we can see that part of correct word-boundary information is retained (Figure \ref{fig:random_matrix} a vs. c). However, it stills hinders the performance in terms of COMET (0.42) and BLEURT (0.20). For the robustness, we find the \textsc{Umst} can still deliver robust performance when injecting noise parser input (Random). Note that we keep the correctness of $\mathcal{A}_w$ and replace the $\mathcal{A}_p$ with the random version of $\mathcal{A}_p$. It obtains 0.14 BLEU gains compared with \textsc{Umst} w/o inter-group, though it still underperforms by the correct parser. 
    
    \paragraph{Quantitative Examples}
    As we discussed in \Cref{sec:intro}, the standard Transformer tends to generate similar and dispersive attention distributions. This can be observed in Figure \ref{fig:attn_map} (above), where attention maps of several sentences in the same batch share a similar distribution, which is not what we expected. For our model, the attention distributions tend to be more diverse and show distinct word boundaries, which follows their own linguistic characteristics. The results are shown in Figure \ref{fig:attn_map} (below). For example, given a sentence ``Intersex children pose ethical dilemma.'' whose BPE-level partition is ``Inter@@ sex children pose ethical dil@@ em@@ ma.'', the areas composed of ``Inter@@'', ``sex'' and composed of ``dil@@'', ``em@@'', ``ma'' show darker color, which indicates strong relationship within the word. Similar phenomena can be observed in the other three cases. More results please refer to \cref{sec:appendix_exp}.

    \section{Related Work}
    \label{sec:related_work}

    Transformer serve as the primary backbone for natural language processing tasks, such as machine translation \cite{vaswani2017attention}, language modeling \cite{baevski2019adaptive,dai-etal-2019-transformer}, pretraining \cite{devlin-etal-2019-bert,brown2020gpt3}, speech processing \cite{gulati2020conformer}. Learning deep Transformer models \cite{wang-etal-2019-learning,li-etal-2020-shallow,brown2020gpt3,li2021lightweight} or improving the parameter efficiency \cite{bai2018deq,li-etal-2022-ode,mehta2020delight} are effective to gain more performance gains. However, they only encode single-scale features instead of multiscale features that are widely used in computer vision.

    \paragraph{Computer vision}
    Multiscale modeling has been witnessed in the computer vision area since the emergence of pyramid feature network \cite{lin2017feature}, which is the main backbone for object-detection, yielding the state-of-the-art performances compared with standard ResNet \cite{he2016deep}. More recently, Vision Transformer (ViT) shows strong ability to handle vision tasks compared with convolutional networks when the data is sufficient. It divides the input feature into several patches and recombines them in a sequential way without architecture modification. The success of ViT attract lots of researchers to design more effective backbones upon it. It is a natural idea to enable ViT to capture patterns at different scales, such as incorporating both local and global patterns \cite{Vaswani-2021-SLSA, Yang-2021-Focalsan,Wu-2021-cvt,Chen-2021-RegionVit,Chu-2021-twins,liu2021swin}, multiscale vision Transformers \cite{fan2021MViT,li2021improvedMViT,wang2021pvt}.

    \paragraph{Natural language processing}
    However, directly transferring the above techniques to natural language processing tasks is nontrivial. A potential explanation is that vision features are better at constructed multi-granularity representations as the consecutive pixels rather than discrete text inputs. As an alternative, researchers redesign the components in Transformer at different granularities. Concretely, \citet{fan2021multiscale} captured locally fine-grained information via restricting the receive filed of heads. Local pattern modeling has been a hot topic, recently. And several works \cite{hao-etal-2019-multi,Zhao-2019-MUSE} have shown local information is indeed helpful as adjacent words are correlated in similar semantic spaces. \citet{wu-etal-2018-phrase} developed a phrase-level self-attention mechanism to improve the computation efficiency. \citet{wei-etal-2020-multiscale} associated the decoder with the encoder with multi-granular dependencies in different space-scales. Another emerging thread of research is to incorporate various input features via combing the sentence-piece and BPE \cite{wu2020mixed} or several BPE vocabularies in different sizes \cite{morishita-etal-2018-improving}.

    \section{Conclusions}
    In this work, we revisit the design of multiscale Transformer for sequence generation. First, we partition scales into sub-words, words and phrases from the linguistic unit perspective. Upon this, we establish interactions among different scales to incorporate the word boundaries and phrase-level information to compensate for the global interactions. Extensive results on two benchmarks, machine translation and abstractive summarization, show the superiority of our method. As another bonus, \textsc{Umst} provides an opportunity of incorporating the prior knowledge into Transformer, which may shed lights on the subsequent researchers. Our code is available at \url{https://github.com/libeineu/UMST}.

    \section*{Acknowledgments}
    This work was supported in part by the National Science Foundation of China (Nos. 61876035 and 61732005), the China HTRD Center Project (No. 2020AAA0107904) and Yunnan Provincial Major Science and Technology Special Plan Projects (Nos. 202002AD080001 and 202103AA080015). The authors would like to thank anonymous reviewers for their valuable comments. And thank Yufan Jiang for his helpful advice to improve the paper.

\nocite{langley00}

\bibliography{multiscale_Transformer}

\begin{thebibliography}{61}
\providecommand{\natexlab}[1]{#1}
\providecommand{\url}[1]{\texttt{#1}}
\expandafter\ifx\csname urlstyle\endcsname\relax
  \providecommand{\doi}[1]{doi: #1}\else
  \providecommand{\doi}{doi: \begingroup \urlstyle{rm}\Url}\fi

\bibitem[Baevski \& Auli(2019)Baevski and Auli]{baevski2019adaptive}
Baevski, A. and Auli, M.
\newblock Adaptive input representations for neural language modeling.
\newblock In \emph{7th International Conference on Learning Representations,
  {ICLR} 2019, New Orleans, LA, USA, May 6-9, 2019}. OpenReview.net, 2019.
\newblock URL \url{https://openreview.net/forum?id=ByxZX20qFQ}.

\bibitem[Bai et~al.(2019)Bai, Kolter, and Koltun]{bai2018deq}
Bai, S., Kolter, J.~Z., and Koltun, V.
\newblock Deep equilibrium models.
\newblock In Wallach, H.~M., Larochelle, H., Beygelzimer, A.,
  d'Alch{\'{e}}{-}Buc, F., Fox, E.~B., and Garnett, R. (eds.), \emph{Advances
  in Neural Information Processing Systems 32: Annual Conference on Neural
  Information Processing Systems 2019, NeurIPS 2019, December 8-14, 2019,
  Vancouver, BC, Canada}, pp.\  688--699, 2019.
\newblock URL
  \url{https://proceedings.neurips.cc/paper/2019/hash/01386bd6d8e091c2ab4c7c7de644d37b-Abstract.html}.

\bibitem[Brown et~al.(2020)Brown, Mann, Ryder, Subbiah, Kaplan, Dhariwal,
  Neelakantan, Shyam, Sastry, Askell, Agarwal, Herbert{-}Voss, Krueger,
  Henighan, Child, Ramesh, Ziegler, Wu, Winter, Hesse, Chen, Sigler, Litwin,
  Gray, Chess, Clark, Berner, McCandlish, Radford, Sutskever, and
  Amodei]{brown2020gpt3}
Brown, T.~B., Mann, B., Ryder, N., Subbiah, M., Kaplan, J., Dhariwal, P.,
  Neelakantan, A., Shyam, P., Sastry, G., Askell, A., Agarwal, S.,
  Herbert{-}Voss, A., Krueger, G., Henighan, T., Child, R., Ramesh, A.,
  Ziegler, D.~M., Wu, J., Winter, C., Hesse, C., Chen, M., Sigler, E., Litwin,
  M., Gray, S., Chess, B., Clark, J., Berner, C., McCandlish, S., Radford, A.,
  Sutskever, I., and Amodei, D.
\newblock Language models are few-shot learners.
\newblock In Larochelle, H., Ranzato, M., Hadsell, R., Balcan, M., and Lin, H.
  (eds.), \emph{Advances in Neural Information Processing Systems 33: Annual
  Conference on Neural Information Processing Systems 2020, NeurIPS 2020,
  December 6-12, 2020, virtual}, 2020.

\bibitem[Carion et~al.(2020)Carion, Massa, Synnaeve, Usunier, Kirillov, and
  Zagoruyko]{carion2020DETR}
Carion, N., Massa, F., Synnaeve, G., Usunier, N., Kirillov, A., and Zagoruyko,
  S.
\newblock End-to-end object detection with transformers.
\newblock In Vedaldi, A., Bischof, H., Brox, T., and Frahm, J. (eds.),
  \emph{Computer Vision - {ECCV} 2020 - 16th European Conference, Glasgow, UK,
  August 23-28, 2020, Proceedings, Part {I}}, volume 12346 of \emph{Lecture
  Notes in Computer Science}, pp.\  213--229. Springer, 2020.

\bibitem[Chen et~al.(2021)Chen, Panda, and Fan]{Chen-2021-RegionVit}
Chen, C., Panda, R., and Fan, Q.
\newblock Regionvit: Regional-to-local attention for vision transformers.
\newblock \emph{CoRR}, abs/2106.02689, 2021.
\newblock URL \url{https://arxiv.org/abs/2106.02689}.

\bibitem[Chu et~al.(2021)Chu, Tian, Wang, Zhang, Ren, Wei, Xia, and
  Shen]{Chu-2021-twins}
Chu, X., Tian, Z., Wang, Y., Zhang, B., Ren, H., Wei, X., Xia, H., and Shen, C.
\newblock Twins: Revisiting spatial attention design in vision transformers.
\newblock \emph{CoRR}, abs/2104.13840, 2021.
\newblock URL \url{https://arxiv.org/abs/2104.13840}.

\bibitem[Dai et~al.(2019)Dai, Yang, Yang, Carbonell, Le, and
  Salakhutdinov]{dai-etal-2019-transformer}
Dai, Z., Yang, Z., Yang, Y., Carbonell, J., Le, Q., and Salakhutdinov, R.
\newblock Transformer-{XL}: Attentive language models beyond a fixed-length
  context.
\newblock In \emph{Proceedings of the 57th Annual Meeting of the Association
  for Computational Linguistics}, pp.\  2978--2988, Florence, Italy, July 2019.
  Association for Computational Linguistics.
\newblock \doi{10.18653/v1/P19-1285}.
\newblock URL \url{https://aclanthology.org/P19-1285}.

\bibitem[Dehghani et~al.(2019)Dehghani, Gouws, Vinyals, Uszkoreit, and
  Kaiser]{Dehghani2019universal}
Dehghani, M., Gouws, S., Vinyals, O., Uszkoreit, J., and Kaiser, L.
\newblock Universal transformers.
\newblock In \emph{7th International Conference on Learning Representations,
  {ICLR} 2019, New Orleans, LA, USA, May 6-9, 2019}. OpenReview.net, 2019.
\newblock URL \url{https://openreview.net/forum?id=HyzdRiR9Y7}.

\bibitem[Devlin et~al.(2019)Devlin, Chang, Lee, and
  Toutanova]{devlin-etal-2019-bert}
Devlin, J., Chang, M.-W., Lee, K., and Toutanova, K.
\newblock {BERT}: Pre-training of deep bidirectional transformers for language
  understanding.
\newblock In \emph{Proceedings of the 2019 Conference of the North {A}merican
  Chapter of the Association for Computational Linguistics: Human Language
  Technologies, Volume 1 (Long and Short Papers)}, pp.\  4171--4186,
  Minneapolis, Minnesota, June 2019. Association for Computational Linguistics.
\newblock \doi{10.18653/v1/N19-1423}.
\newblock URL \url{https://aclanthology.org/N19-1423}.

\bibitem[Dosovitskiy et~al.(2021)Dosovitskiy, Beyer, Kolesnikov, Weissenborn,
  Zhai, Unterthiner, Dehghani, Minderer, Heigold, Gelly, Uszkoreit, and
  Houlsby]{dosovitskiy2020ViT}
Dosovitskiy, A., Beyer, L., Kolesnikov, A., Weissenborn, D., Zhai, X.,
  Unterthiner, T., Dehghani, M., Minderer, M., Heigold, G., Gelly, S.,
  Uszkoreit, J., and Houlsby, N.
\newblock An image is worth 16x16 words: Transformers for image recognition at
  scale.
\newblock In \emph{9th International Conference on Learning Representations,
  {ICLR} 2021, Virtual Event, Austria, May 3-7, 2021}. OpenReview.net, 2021.
\newblock URL \url{https://openreview.net/forum?id=YicbFdNTTy}.

\bibitem[Fan et~al.(2021{\natexlab{a}})Fan, Xiong, Mangalam, Li, Yan, Malik,
  and Feichtenhofer]{fan2021MViT}
Fan, H., Xiong, B., Mangalam, K., Li, Y., Yan, Z., Malik, J., and
  Feichtenhofer, C.
\newblock Multiscale vision transformers.
\newblock \emph{CoRR}, abs/2104.11227, 2021{\natexlab{a}}.
\newblock URL \url{https://arxiv.org/abs/2104.11227}.

\bibitem[Fan et~al.(2021{\natexlab{b}})Fan, Xiong, Mangalam, Li, Yan, Malik,
  and Feichtenhofer]{fan2021multiscale}
Fan, H., Xiong, B., Mangalam, K., Li, Y., Yan, Z., Malik, J., and
  Feichtenhofer, C.
\newblock Multiscale vision transformers.
\newblock \emph{arXiv preprint arXiv:2104.11227}, 2021{\natexlab{b}}.

\bibitem[Fan et~al.(2021{\natexlab{c}})Fan, Gong, Liu, Wei, Wang, Jiao, Duan,
  Zhang, and Huang]{fan-etal-2021-mask}
Fan, Z., Gong, Y., Liu, D., Wei, Z., Wang, S., Jiao, J., Duan, N., Zhang, R.,
  and Huang, X.
\newblock Mask attention networks: Rethinking and strengthen transformer.
\newblock In \emph{Proceedings of the 2021 Conference of the North American
  Chapter of the Association for Computational Linguistics: Human Language
  Technologies}, pp.\  1692--1701, Online, June 2021{\natexlab{c}}. Association
  for Computational Linguistics.
\newblock \doi{10.18653/v1/2021.naacl-main.135}.
\newblock URL \url{https://aclanthology.org/2021.naacl-main.135}.

\bibitem[Gehrmann et~al.(2018)Gehrmann, Deng, and
  Rush]{gehrmann-etal-2018-bottom}
Gehrmann, S., Deng, Y., and Rush, A.
\newblock Bottom-up abstractive summarization.
\newblock In \emph{Proceedings of the 2018 Conference on Empirical Methods in
  Natural Language Processing}, pp.\  4098--4109, Brussels, Belgium,
  October-November 2018.

\bibitem[Gulati et~al.(2020)Gulati, Qin, Chiu, Parmar, Zhang, Yu, Han, Wang,
  Zhang, Wu, and Pang]{gulati2020conformer}
Gulati, A., Qin, J., Chiu, C., Parmar, N., Zhang, Y., Yu, J., Han, W., Wang,
  S., Zhang, Z., Wu, Y., and Pang, R.
\newblock Conformer: Convolution-augmented transformer for speech recognition.
\newblock In Meng, H., Xu, B., and Zheng, T.~F. (eds.), \emph{Interspeech 2020,
  21st Annual Conference of the International Speech Communication Association,
  Virtual Event, Shanghai, China, 25-29 October 2020}, pp.\  5036--5040.
  {ISCA}, 2020.
\newblock URL \url{https://doi.org/10.21437/Interspeech.2020-3015}.

\bibitem[Guo et~al.(2020)Guo, Qiu, Liu, Xue, and Zhang]{guo-2020-msan}
Guo, Q., Qiu, X., Liu, P., Xue, X., and Zhang, Z.
\newblock Multi-scale self-attention for text classification.
\newblock In \emph{The Thirty-Fourth {AAAI} Conference on Artificial
  Intelligence, {AAAI} 2020, The Thirty-Second Innovative Applications of
  Artificial Intelligence Conference, {IAAI} 2020, The Tenth {AAAI} Symposium
  on Educational Advances in Artificial Intelligence, {EAAI} 2020, New York,
  NY, USA, February 7-12, 2020}, pp.\  7847--7854. {AAAI} Press, 2020.
\newblock URL \url{https://aaai.org/ojs/index.php/AAAI/article/view/6290}.

\bibitem[Han et~al.(2021)Han, Xiao, Wu, Guo, Xu, and Wang]{han2021transformer}
Han, K., Xiao, A., Wu, E., Guo, J., Xu, C., and Wang, Y.
\newblock Transformer in transformer.
\newblock \emph{arXiv preprint arXiv:2103.00112}, 2021.

\bibitem[Hao et~al.(2019)Hao, Wang, Shi, Zhang, and Tu]{hao-etal-2019-multi}
Hao, J., Wang, X., Shi, S., Zhang, J., and Tu, Z.
\newblock Multi-granularity self-attention for neural machine translation.
\newblock In \emph{Proceedings of the 2019 Conference on Empirical Methods in
  Natural Language Processing and the 9th International Joint Conference on
  Natural Language Processing (EMNLP-IJCNLP)}, pp.\  887--897, Hong Kong,
  China, November 2019. Association for Computational Linguistics.
\newblock \doi{10.18653/v1/D19-1082}.
\newblock URL \url{https://aclanthology.org/D19-1082}.

\bibitem[He et~al.(2016)He, Zhang, Ren, and Sun]{he2016deep}
He, K., Zhang, X., Ren, S., and Sun, J.
\newblock Deep residual learning for image recognition.
\newblock In \emph{2016 {IEEE} Conference on Computer Vision and Pattern
  Recognition, {CVPR} 2016, Las Vegas, NV, USA, June 27-30, 2016}, pp.\
  770--778. {IEEE} Computer Society, 2016.
\newblock \doi{10.1109/CVPR.2016.90}.
\newblock URL \url{https://doi.org/10.1109/CVPR.2016.90}.

\bibitem[Hermann et~al.(2015)Hermann, Kocisk{\'{y}}, Grefenstette, Espeholt,
  Kay, Suleyman, and Blunsom]{hermann2015teaching}
Hermann, K.~M., Kocisk{\'{y}}, T., Grefenstette, E., Espeholt, L., Kay, W.,
  Suleyman, M., and Blunsom, P.
\newblock Teaching machines to read and comprehend.
\newblock In Cortes, C., Lawrence, N.~D., Lee, D.~D., Sugiyama, M., and
  Garnett, R. (eds.), \emph{Advances in Neural Information Processing Systems
  28: Annual Conference on Neural Information Processing Systems 2015, December
  7-12, 2015, Montreal, Quebec, Canada}, pp.\  1693--1701, 2015.
\newblock URL
  \url{https://proceedings.neurips.cc/paper/2015/hash/afdec7005cc9f14302cd0474fd0f3c96-Abstract.html}.

\bibitem[Kasai et~al.(2020)Kasai, Cross, Ghazvininejad, and Gu]{kasai2020disco}
Kasai, J., Cross, J., Ghazvininejad, M., and Gu, J.
\newblock Non-autoregressive machine translation with disentangled context
  transformer.
\newblock In \emph{Proceedings of the 37th International Conference on Machine
  Learning, {ICML} 2020, 13-18 July 2020, Virtual Event}, volume 119 of
  \emph{Proceedings of Machine Learning Research}, pp.\  5144--5155. {PMLR},
  2020.
\newblock URL \url{http://proceedings.mlr.press/v119/kasai20a.html}.

\bibitem[Kingma \& Ba(2015)Kingma and Ba]{kingma2014adam}
Kingma, D.~P. and Ba, J.
\newblock Adam: {A} method for stochastic optimization.
\newblock In Bengio, Y. and LeCun, Y. (eds.), \emph{3rd International
  Conference on Learning Representations, {ICLR} 2015, San Diego, CA, USA, May
  7-9, 2015, Conference Track Proceedings}, 2015.
\newblock URL \url{http://arxiv.org/abs/1412.6980}.

\bibitem[Kipf \& Welling(2017)Kipf and Welling]{Kipf2017gcn}
Kipf, T.~N. and Welling, M.
\newblock Semi-supervised classification with graph convolutional networks.
\newblock In \emph{5th International Conference on Learning Representations,
  {ICLR} 2017, Toulon, France, April 24-26, 2017, Conference Track
  Proceedings}. OpenReview.net, 2017.
\newblock URL \url{https://openreview.net/forum?id=SJU4ayYgl}.

\bibitem[Kitaev \& Klein(2018)Kitaev and Klein]{kitaev-klein-2018-constituency}
Kitaev, N. and Klein, D.
\newblock Constituency parsing with a self-attentive encoder.
\newblock In \emph{Proceedings of the 56th Annual Meeting of the Association
  for Computational Linguistics (Volume 1: Long Papers)}, pp.\  2676--2686,
  Melbourne, Australia, July 2018. Association for Computational Linguistics.
\newblock \doi{10.18653/v1/P18-1249}.
\newblock URL \url{https://www.aclweb.org/anthology/P18-1249}.

\bibitem[Kudo \& Richardson(2018)Kudo and
  Richardson]{kudo-richardson-2018-sentencepiece}
Kudo, T. and Richardson, J.
\newblock {S}entence{P}iece: A simple and language independent subword
  tokenizer and detokenizer for neural text processing.
\newblock In \emph{Proceedings of the 2018 Conference on Empirical Methods in
  Natural Language Processing: System Demonstrations}, pp.\  66--71, Brussels,
  Belgium, November 2018. Association for Computational Linguistics.
\newblock \doi{10.18653/v1/D18-2012}.
\newblock URL \url{https://aclanthology.org/D18-2012}.

\bibitem[Lee et~al.(2018)Lee, Mansimov, and Cho]{lee-etal-2018-deterministic}
Lee, J., Mansimov, E., and Cho, K.
\newblock Deterministic non-autoregressive neural sequence modeling by
  iterative refinement.
\newblock In \emph{Proceedings of the 2018 Conference on Empirical Methods in
  Natural Language Processing}, pp.\  1173--1182, Brussels, Belgium,
  October-November 2018. Association for Computational Linguistics.
\newblock \doi{10.18653/v1/D18-1149}.
\newblock URL \url{https://aclanthology.org/D18-1149}.

\bibitem[Lei~Ba et~al.(2016)Lei~Ba, Kiros, and Hinton]{lei2016layer}
Lei~Ba, J., Kiros, J.~R., and Hinton, G.~E.
\newblock Layer normalization.
\newblock \emph{ArXiv preprint}, abs/1607.06450, 2016.
\newblock URL \url{https://arxiv.org/abs/1607.06450}.

\bibitem[Li et~al.(2020)Li, Wang, Liu, Jiang, Du, Xiao, Wang, and
  Zhu]{li-etal-2020-shallow}
Li, B., Wang, Z., Liu, H., Jiang, Y., Du, Q., Xiao, T., Wang, H., and Zhu, J.
\newblock Shallow-to-deep training for neural machine translation.
\newblock In \emph{Proceedings of the 2020 Conference on Empirical Methods in
  Natural Language Processing (EMNLP)}, pp.\  995--1005, Online, November 2020.
  Association for Computational Linguistics.
\newblock \doi{10.18653/v1/2020.emnlp-main.72}.
\newblock URL \url{https://aclanthology.org/2020.emnlp-main.72}.

\bibitem[Li et~al.(2021{\natexlab{a}})Li, Wang, Liu, Du, Xiao, Zhang, and
  Zhu]{li2021lightweight}
Li, B., Wang, Z., Liu, H., Du, Q., Xiao, T., Zhang, C., and Zhu, J.
\newblock Learning light-weight translation models from deep transformer.
\newblock In \emph{Thirty-Fifth {AAAI} Conference on Artificial Intelligence,
  {AAAI} 2021, Thirty-Third Conference on Innovative Applications of Artificial
  Intelligence, {IAAI} 2021, The Eleventh Symposium on Educational Advances in
  Artificial Intelligence, {EAAI} 2021, Virtual Event, February 2-9, 2021},
  pp.\  13217--13225. {AAAI} Press, 2021{\natexlab{a}}.
\newblock URL \url{https://ojs.aaai.org/index.php/AAAI/article/view/17561}.

\bibitem[Li et~al.(2022)Li, Du, Zhou, Jing, Zhou, Zeng, Xiao, Zhu, Liu, and
  Zhang]{li-etal-2022-ode}
Li, B., Du, Q., Zhou, T., Jing, Y., Zhou, S., Zeng, X., Xiao, T., Zhu, J., Liu,
  X., and Zhang, M.
\newblock {ODE} transformer: An ordinary differential equation-inspired model
  for sequence generation.
\newblock In \emph{Proceedings of the 60th Annual Meeting of the Association
  for Computational Linguistics (Volume 1: Long Papers)}, pp.\  8335--8351,
  Dublin, Ireland, May 2022. Association for Computational Linguistics.
\newblock \doi{10.18653/v1/2022.acl-long.571}.
\newblock URL \url{https://aclanthology.org/2022.acl-long.571}.

\bibitem[Li et~al.(2021{\natexlab{b}})Li, Wu, Fan, Mangalam, Xiong, Malik, and
  Feichtenhofer]{li2021improvedMViT}
Li, Y., Wu, C., Fan, H., Mangalam, K., Xiong, B., Malik, J., and Feichtenhofer,
  C.
\newblock Improved multiscale vision transformers for classification and
  detection.
\newblock \emph{CoRR}, abs/2112.01526, 2021{\natexlab{b}}.
\newblock URL \url{https://arxiv.org/abs/2112.01526}.

\bibitem[Lin(2004)]{lin-2004-rouge}
Lin, C.-Y.
\newblock {ROUGE}: A package for automatic evaluation of summaries.
\newblock In \emph{Text Summarization Branches Out}, pp.\  74--81, Barcelona,
  Spain, 2004. Association for Computational Linguistics.
\newblock URL \url{https://aclanthology.org/W04-1013}.

\bibitem[Lin et~al.(2017)Lin, Doll{\'a}r, Girshick, He, Hariharan, and
  Belongie]{lin2017feature}
Lin, T.-Y., Doll{\'a}r, P., Girshick, R., He, K., Hariharan, B., and Belongie,
  S.
\newblock Feature pyramid networks for object detection.
\newblock In \emph{Proceedings of the IEEE conference on computer vision and
  pattern recognition}, pp.\  2117--2125, 2017.

\bibitem[Liu et~al.(2020{\natexlab{a}})Liu, Wang, Wong, Ding, Chao, and
  Tu]{liu2020understanding}
Liu, X., Wang, L., Wong, D.~F., Ding, L., Chao, L.~S., and Tu, Z.
\newblock Understanding and improving encoder layer fusion in
  sequence-to-sequence learning.
\newblock \emph{ArXiv preprint}, abs/2012.14768, 2020{\natexlab{a}}.
\newblock URL \url{https://arxiv.org/abs/2012.14768}.

\bibitem[Liu et~al.(2020{\natexlab{b}})Liu, Gu, Goyal, Li, Edunov,
  Ghazvininejad, Lewis, and Zettlemoyer]{liu-etal-2020-multilingual-denoising}
Liu, Y., Gu, J., Goyal, N., Li, X., Edunov, S., Ghazvininejad, M., Lewis, M.,
  and Zettlemoyer, L.
\newblock Multilingual denoising pre-training for neural machine translation.
\newblock \emph{Transactions of the Association for Computational Linguistics},
  8:\penalty0 726--742, 2020{\natexlab{b}}.
\newblock \doi{10.1162/tacl_a_00343}.
\newblock URL \url{https://aclanthology.org/2020.tacl-1.47}.

\bibitem[Liu et~al.(2021)Liu, Lin, Cao, Hu, Wei, Zhang, Lin, and
  Guo]{liu2021swin}
Liu, Z., Lin, Y., Cao, Y., Hu, H., Wei, Y., Zhang, Z., Lin, S., and Guo, B.
\newblock Swin transformer: Hierarchical vision transformer using shifted
  windows.
\newblock \emph{CoRR}, abs/2103.14030, 2021.
\newblock URL \url{https://arxiv.org/abs/2103.14030}.

\bibitem[Mehta et~al.(2020)Mehta, Ghazvininejad, Iyer, Zettlemoyer, and
  Hajishirzi]{mehta2020delight}
Mehta, S., Ghazvininejad, M., Iyer, S., Zettlemoyer, L., and Hajishirzi, H.
\newblock Delight: Very deep and light-weight transformer.
\newblock \emph{ArXiv preprint}, abs/2008.00623, 2020.
\newblock URL \url{https://arxiv.org/abs/2008.00623}.

\bibitem[Morishita et~al.(2018)Morishita, Suzuki, and
  Nagata]{morishita-etal-2018-improving}
Morishita, M., Suzuki, J., and Nagata, M.
\newblock Improving neural machine translation by incorporating hierarchical
  subword features.
\newblock In \emph{Proceedings of the 27th International Conference on
  Computational Linguistics}, pp.\  618--629, Santa Fe, New Mexico, USA, August
  2018. Association for Computational Linguistics.
\newblock URL \url{https://aclanthology.org/C18-1052}.

\bibitem[Nallapati et~al.(2016)Nallapati, Zhou, dos Santos,
  G{\"{u}}l{\c{c}}ehre, and Xiang]{nallapati-etal-2016-abstractive}
Nallapati, R., Zhou, B., dos Santos, C.~N., G{\"{u}}l{\c{c}}ehre, {\c{C}}., and
  Xiang, B.
\newblock Abstractive text summarization using sequence-to-sequence rnns and
  beyond.
\newblock In Goldberg, Y. and Riezler, S. (eds.), \emph{Proceedings of the 20th
  {SIGNLL} Conference on Computational Natural Language Learning, CoNLL 2016,
  Berlin, Germany, August 11-12, 2016}, pp.\  280--290. {ACL}, 2016.

\bibitem[Ott et~al.(2018)Ott, Edunov, Grangier, and Auli]{ott2018scaling}
Ott, M., Edunov, S., Grangier, D., and Auli, M.
\newblock Scaling neural machine translation.
\newblock In \emph{Proceedings of the Third Conference on Machine Translation:
  Research Papers}, pp.\  1--9, Brussels, Belgium, 2018. Association for
  Computational Linguistics.
\newblock \doi{10.18653/v1/W18-6301}.
\newblock URL \url{https://aclanthology.org/W18-6301}.

\bibitem[Ott et~al.(2019)Ott, Edunov, Baevski, Fan, Gross, Ng, Grangier, and
  Auli]{ott2019fairseq}
Ott, M., Edunov, S., Baevski, A., Fan, A., Gross, S., Ng, N., Grangier, D., and
  Auli, M.
\newblock fairseq: A fast, extensible toolkit for sequence modeling.
\newblock In \emph{Proceedings of the 2019 Conference of the North {A}merican
  Chapter of the Association for Computational Linguistics (Demonstrations)},
  pp.\  48--53, Minneapolis, Minnesota, 2019. Association for Computational
  Linguistics.
\newblock \doi{10.18653/v1/N19-4009}.
\newblock URL \url{https://aclanthology.org/N19-4009}.

\bibitem[Qi et~al.(2018)Qi, Dozat, Zhang, and Manning]{qi2018universal}
Qi, P., Dozat, T., Zhang, Y., and Manning, C.~D.
\newblock Universal dependency parsing from scratch.
\newblock In \emph{Proceedings of the {CoNLL} 2018 Shared Task: Multilingual
  Parsing from Raw Text to Universal Dependencies}, pp.\  160--170, Brussels,
  Belgium, October 2018. Association for Computational Linguistics.
\newblock URL \url{https://nlp.stanford.edu/pubs/qi2018universal.pdf}.

\bibitem[Rei et~al.(2020)Rei, Stewart, Farinha, and Lavie]{rei-etal-2020-comet}
Rei, R., Stewart, C., Farinha, A.~C., and Lavie, A.
\newblock {COMET}: A neural framework for {MT} evaluation.
\newblock In \emph{Proceedings of the 2020 Conference on Empirical Methods in
  Natural Language Processing (EMNLP)}, pp.\  2685--2702, Online, November
  2020. Association for Computational Linguistics.
\newblock \doi{10.18653/v1/2020.emnlp-main.213}.
\newblock URL \url{https://aclanthology.org/2020.emnlp-main.213}.

\bibitem[Sellam et~al.(2020)Sellam, Das, and Parikh]{sellam-etal-2020-bleurt}
Sellam, T., Das, D., and Parikh, A.
\newblock {BLEURT}: Learning robust metrics for text generation.
\newblock In \emph{Proceedings of the 58th Annual Meeting of the Association
  for Computational Linguistics}, pp.\  7881--7892, Online, July 2020.
  Association for Computational Linguistics.
\newblock \doi{10.18653/v1/2020.acl-main.704}.
\newblock URL \url{https://aclanthology.org/2020.acl-main.704}.

\bibitem[Sennrich et~al.(2016)Sennrich, Haddow, and
  Birch]{sennrich-subword-neural}
Sennrich, R., Haddow, B., and Birch, A.
\newblock Neural machine translation of rare words with subword units.
\newblock In \emph{Proceedings of the 54th Annual Meeting of the Association
  for Computational Linguistics (Volume 1: Long Papers)}, pp.\  1715--1725,
  Berlin, Germany, 2016. Association for Computational Linguistics.
\newblock \doi{10.18653/v1/P16-1162}.
\newblock URL \url{https://aclanthology.org/P16-1162}.

\bibitem[Shaw et~al.(2018)Shaw, Uszkoreit, and Vaswani]{shaw-etal-2018-self}
Shaw, P., Uszkoreit, J., and Vaswani, A.
\newblock Self-attention with relative position representations.
\newblock In \emph{Proceedings of the 2018 Conference of the North {A}merican
  Chapter of the Association for Computational Linguistics: Human Language
  Technologies, Volume 2 (Short Papers)}, pp.\  464--468, New Orleans,
  Louisiana, 2018. Association for Computational Linguistics.
\newblock \doi{10.18653/v1/N18-2074}.
\newblock URL \url{https://aclanthology.org/N18-2074}.

\bibitem[Sutskever et~al.(2014)Sutskever, Vinyals, and
  Le]{sutskever2014sequence}
Sutskever, I., Vinyals, O., and Le, Q.~V.
\newblock Sequence to sequence learning with neural networks.
\newblock In Ghahramani, Z., Welling, M., Cortes, C., Lawrence, N.~D., and
  Weinberger, K.~Q. (eds.), \emph{Advances in Neural Information Processing
  Systems 27: Annual Conference on Neural Information Processing Systems 2014,
  December 8-13 2014, Montreal, Quebec, Canada}, pp.\  3104--3112, 2014.
\newblock URL
  \url{https://proceedings.neurips.cc/paper/2014/hash/a14ac55a4f27472c5d894ec1c3c743d2-Abstract.html}.

\bibitem[Vaswani et~al.(2017)Vaswani, Shazeer, Parmar, Uszkoreit, Jones, Gomez,
  Kaiser, and Polosukhin]{vaswani2017attention}
Vaswani, A., Shazeer, N., Parmar, N., Uszkoreit, J., Jones, L., Gomez, A.~N.,
  Kaiser, L., and Polosukhin, I.
\newblock Attention is all you need.
\newblock In Guyon, I., von Luxburg, U., Bengio, S., Wallach, H.~M., Fergus,
  R., Vishwanathan, S. V.~N., and Garnett, R. (eds.), \emph{Advances in Neural
  Information Processing Systems 30: Annual Conference on Neural Information
  Processing Systems 2017, December 4-9, 2017, Long Beach, CA, {USA}}, pp.\
  5998--6008, 2017.
\newblock URL
  \url{https://proceedings.neurips.cc/paper/2017/hash/3f5ee243547dee91fbd053c1c4a845aa-Abstract.html}.

\bibitem[Vaswani et~al.(2021)Vaswani, Ramachandran, Srinivas, Parmar, Hechtman,
  and Shlens]{Vaswani-2021-SLSA}
Vaswani, A., Ramachandran, P., Srinivas, A., Parmar, N., Hechtman, B.~A., and
  Shlens, J.
\newblock Scaling local self-attention for parameter efficient visual
  backbones.
\newblock In \emph{{IEEE} Conference on Computer Vision and Pattern
  Recognition, {CVPR} 2021, virtual, June 19-25, 2021}, pp.\  12894--12904.
  Computer Vision Foundation / {IEEE}, 2021.
\newblock URL
  \url{https://openaccess.thecvf.com/content/CVPR2021/html/Vaswani\_Scaling\_Local\_Self-Attention\_}.

\bibitem[Velickovic et~al.(2018)Velickovic, Cucurull, Casanova, Romero,
  Li{\`{o}}, and Bengio]{Petar2018GAT}
Velickovic, P., Cucurull, G., Casanova, A., Romero, A., Li{\`{o}}, P., and
  Bengio, Y.
\newblock Graph attention networks.
\newblock In \emph{6th International Conference on Learning Representations,
  {ICLR} 2018, Vancouver, BC, Canada, April 30 - May 3, 2018, Conference Track
  Proceedings}. OpenReview.net, 2018.
\newblock URL \url{https://openreview.net/forum?id=rJXMpikCZ}.

\bibitem[Wang et~al.(2019)Wang, Li, Xiao, Zhu, Li, Wong, and
  Chao]{wang-etal-2019-learning}
Wang, Q., Li, B., Xiao, T., Zhu, J., Li, C., Wong, D.~F., and Chao, L.~S.
\newblock Learning deep transformer models for machine translation.
\newblock In \emph{Proceedings of the 57th Annual Meeting of the Association
  for Computational Linguistics}, pp.\  1810--1822, Florence, Italy, 2019.
  Association for Computational Linguistics.
\newblock \doi{10.18653/v1/P19-1176}.
\newblock URL \url{https://aclanthology.org/P19-1176}.

\bibitem[Wang et~al.(2021)Wang, Xie, Li, Fan, Song, Liang, Lu, Luo, and
  Shao]{wang2021pvt}
Wang, W., Xie, E., Li, X., Fan, D., Song, K., Liang, D., Lu, T., Luo, P., and
  Shao, L.
\newblock Pyramid vision transformer: {A} versatile backbone for dense
  prediction without convolutions.
\newblock \emph{CoRR}, abs/2102.12122, 2021.
\newblock URL \url{https://arxiv.org/abs/2102.12122}.

\bibitem[Wei et~al.(2020)Wei, Yu, Hu, Zhang, Weng, and
  Luo]{wei-etal-2020-multiscale}
Wei, X., Yu, H., Hu, Y., Zhang, Y., Weng, R., and Luo, W.
\newblock Multiscale collaborative deep models for neural machine translation.
\newblock In \emph{Proceedings of the 58th Annual Meeting of the Association
  for Computational Linguistics}, pp.\  414--426, Online, 2020. Association for
  Computational Linguistics.
\newblock \doi{10.18653/v1/2020.acl-main.40}.
\newblock URL \url{https://aclanthology.org/2020.acl-main.40}.

\bibitem[Wu et~al.(2019)Wu, Fan, Baevski, Dauphin, and Auli]{wu2019DynamicConv}
Wu, F., Fan, A., Baevski, A., Dauphin, Y.~N., and Auli, M.
\newblock Pay less attention with lightweight and dynamic convolutions.
\newblock In \emph{7th International Conference on Learning Representations,
  {ICLR} 2019, New Orleans, LA, USA, May 6-9, 2019}. OpenReview.net, 2019.
\newblock URL \url{https://openreview.net/forum?id=SkVhlh09tX}.

\bibitem[Wu et~al.(2021)Wu, Xiao, Codella, Liu, Dai, Yuan, and
  Zhang]{Wu-2021-cvt}
Wu, H., Xiao, B., Codella, N., Liu, M., Dai, X., Yuan, L., and Zhang, L.
\newblock Cvt: Introducing convolutions to vision transformers.
\newblock \emph{CoRR}, abs/2103.15808, 2021.
\newblock URL \url{https://arxiv.org/abs/2103.15808}.

\bibitem[Wu et~al.(2020)Wu, Xie, Xia, Fan, Lai, Qin, and Liu]{wu2020mixed}
Wu, L., Xie, S., Xia, Y., Fan, Y., Lai, J., Qin, T., and Liu, T.
\newblock Sequence generation with mixed representations.
\newblock In \emph{Proceedings of the 37th International Conference on Machine
  Learning, {ICML} 2020, 13-18 July 2020, Virtual Event}, volume 119 of
  \emph{Proceedings of Machine Learning Research}, pp.\  10388--10398. {PMLR},
  2020.
\newblock URL \url{http://proceedings.mlr.press/v119/wu20e.html}.

\bibitem[Wu et~al.(2018)Wu, Wang, Liu, and Ma]{wu-etal-2018-phrase}
Wu, W., Wang, H., Liu, T., and Ma, S.
\newblock Phrase-level self-attention networks for universal sentence encoding.
\newblock In \emph{Proceedings of the 2018 Conference on Empirical Methods in
  Natural Language Processing}, pp.\  3729--3738, Brussels, Belgium,
  October-November 2018. Association for Computational Linguistics.
\newblock \doi{10.18653/v1/D18-1408}.
\newblock URL \url{https://aclanthology.org/D18-1408}.

\bibitem[Xu et~al.(2021)Xu, Zhou, Gan, Zheng, and Li]{xu-etal-2021-vocabulary}
Xu, J., Zhou, H., Gan, C., Zheng, Z., and Li, L.
\newblock Vocabulary learning via optimal transport for neural machine
  translation.
\newblock In \emph{Proceedings of the 59th Annual Meeting of the Association
  for Computational Linguistics and the 11th International Joint Conference on
  Natural Language Processing (Volume 1: Long Papers)}, pp.\  7361--7373,
  Online, August 2021. Association for Computational Linguistics.
\newblock \doi{10.18653/v1/2021.acl-long.571}.
\newblock URL \url{https://aclanthology.org/2021.acl-long.571}.

\bibitem[Yang et~al.(2019)Yang, Wang, Wong, Chao, and
  Tu]{yang-etal-2019-convolutional}
Yang, B., Wang, L., Wong, D.~F., Chao, L.~S., and Tu, Z.
\newblock Convolutional self-attention networks.
\newblock In \emph{Proceedings of the 2019 Conference of the North {A}merican
  Chapter of the Association for Computational Linguistics: Human Language
  Technologies, Volume 1 (Long and Short Papers)}, pp.\  4040--4045,
  Minneapolis, Minnesota, June 2019. Association for Computational Linguistics.
\newblock \doi{10.18653/v1/N19-1407}.
\newblock URL \url{https://aclanthology.org/N19-1407}.

\bibitem[Yang et~al.(2021)Yang, Li, Zhang, Dai, Xiao, Yuan, and
  Gao]{Yang-2021-Focalsan}
Yang, J., Li, C., Zhang, P., Dai, X., Xiao, B., Yuan, L., and Gao, J.
\newblock Focal self-attention for local-global interactions in vision
  transformers.
\newblock \emph{CoRR}, abs/2107.00641, 2021.
\newblock URL \url{https://arxiv.org/abs/2107.00641}.

\bibitem[Zhao et~al.(2019)Zhao, Sun, Xu, Zhang, and Luo]{Zhao-2019-MUSE}
Zhao, G., Sun, X., Xu, J., Zhang, Z., and Luo, L.
\newblock {MUSE:} parallel multi-scale attention for sequence to sequence
  learning.
\newblock \emph{CoRR}, abs/1911.09483, 2019.
\newblock URL \url{http://arxiv.org/abs/1911.09483}.

\end{thebibliography}
\bibliographystyle{icml2022}

\newpage
\appendix
\onecolumn
\section{Experimental Setups}
\label{sec:exp_setup}
\subsection{Hyperparameters}
We choose Pre-Norm Transformer as the backbone due to its training stability. All systems were trained via Adam optimizer \cite{kingma2014adam}, where $\beta_1$ and $\beta_2$ were set to 0.9 and 0.997. The learning rate and warmup-step were $2e^{-3}$/16000 and $2e^{-3}$/8000 for the machine translation and abstractive summarization tasks, respectively. Our codebase is built based on \texttt{Fairseq} \cite{ott2019fairseq}, and it would be publicly available soon.

For the machine translation task, we measured the performance in terms of BLEU using beam search strategy, where the beam width was 4 and the length penalty was 0.6. For the abstractive summarization task, we reported the Rouge-1, Rouge-2 and Rouge-L \cite{lin-2004-rouge} for comparisons with previous works.

\subsection{Configuration Details}
We mainly built the model based on Transformer Deep and Big configurations. For the deep model, the hidden size is 512 and the filter size of FFN is 2048. We split the hidden space into 8 pieces for the multi-head attention mechanism. The values of dropout are set to 0.1, and so as the label smoothing. For the big model, the hidden size and the filter size are twice larger compared with the deep model. Note that the residual dropout is 0.3 for big models.

For fair comparisons, we re-implement several strongly related works under the same codebase. Also, we follow the setup in \citet{ott2018scaling}'s work to simulate a 128-gpu batching schema via the gradient accumulation strategy, where the max-token size is 9600 and every 8 steps to update the parameters. Note that \citet{Zhao-2019-MUSE} adopted cosine learning rate schema, we still use the standard decay schedule which is proportionally to the inverse square root of the current step. Thus the experiments listed may slightly different from theirs, but resulting in a fairer comparison.

\section{Additional Experiments and Analysis}
\label{sec:appendix_exp}
\subsection{Comparisons with Previous Works}
As we discussed in \cref{sec:intro}, multiscale feature hierarchies have been successfully applied to convolutional and self-attentional networks, e.g. Transformer-in-Transformer (namely TNT) \cite{han2021transformer}, Multiscale Vision Transformer (namely MViT) \cite{fan2021MViT}. Here, we follow their merits and redesign the Transformer architecture. The details are as follows:
\definecolor{upurple}{RGB}{155,89,182}
\definecolor{ublue}{RGB}{52,152,219}
\definecolor{ured}{RGB}{255,160,112}
\definecolor{udark}{RGB}{77,153,77}
\definecolor{ugreen}{RGB}{46,204,113}
\definecolor{darkgreen}{RGB}{61,145,64}
\definecolor{goldenorange}{RGB}{255,153,18}
\definecolor{oldorange}{RGB}{255,153,0}
\definecolor{violet}{RGB}{238, 130,238}
\definecolor{royalblue}{RGB}{65,102,225}
\definecolor{CYAN}{RGB}{65,102,225}

\definecolor{tiffanyblue}{RGB}{129,216,208}
\definecolor{bangdiblue}{RGB}{0,149,182}
\definecolor{kleinblue}{RGB}{0,47,167}
\definecolor{kabuliblue}{RGB}{26,85,153}

\newlength{\tntwidth}
	    \setlength{\tntwidth}{3em}     
	    \newlength{\tntheight}
	    \setlength{\tntheight}{3em}    
		\newlength{\modulelen}
		\setlength{\modulelen}{3em}    
		
	\tikzstyle{circlenode}=[draw,circle,minimum size=4pt,inner sep=0,thick]
    \tikzstyle{add}=[circle,minimum size=1.2em,inner sep=0]
    \tikzstyle{boardernode}=[draw,minimum height=8em,minimum width=15em,inner sep=0,thick,rounded corners=.2em]
    \tikzstyle{boardernode1}=[draw,minimum height=3em,minimum width=15em,inner sep=0,thick,rounded corners=.2em]
    \tikzstyle{boardernode2}=[draw,minimum height=2em,minimum width=10em,inner sep=0,thick,rounded corners=.2em]
    \tikzstyle{modulenode}=[draw,minimum height=2.4em,minimum width=5em,inner sep=0.3em,thick,rounded corners=.2em,thick]
    \tikzstyle{modulenode1}=[draw,minimum height=1.5em,minimum width=3em,inner sep=0.3em,thick,rounded corners=.2em,thick]
    \tikzstyle{USnode}=[draw,minimum height=1.4em,minimum width=2.4em,inner sep=0em,thick,rounded corners=.2em,thick]

\begin{figure*}[htbp]
    \centering
	\subfigure[TNT transformer] { \label{fig:tnttrans} 
    \centering
	\begin{tikzpicture}
    \node[] (input) at (2.5em,0em) {$X^{L \times L}$};
    \node[] (input2bpe) at (0em,7em) {$X_{bpe}^{L \times L}$};
    \node[] (input2word) at (5em,7em) {$X_{word}^{L^{'} \times L^{'}}$};
    \node[modulenode1, align=center, fill=ured] (module11)at (0em,4em){Identity};
	\node[modulenode1, align=center, fill=violet] (module21)at (5em,4em){DS};
	\draw[thick, ->] (input) -| (module11);
	\draw[thick, ->] (module11) edge (input2bpe);
	\draw[thick, ->] (input) -| (module21);
	\draw[thick, ->] (module21) edge (input2word);
	\node[boardernode1, fill=gray!10] (boarder1) at (2.5em,10em){};
	\node[boardernode2, fill=goldenorange!30] (boarder2) at (2.5em,10em){TNT block};
	\node[] (label) at ([xshift=6em]boarder2.center) {$\times M$};
	\draw[->, thick] (input2bpe) edge ([xshift=-2.5em, yshift=-1em]boarder2);
	\draw[->, thick] (input2word) edge ([xshift=2.5em, yshift=-1em]boarder2);
	\draw[->, thick] ([xshift=-2.5em,yshift=1em]boarder2.center) -- ([xshift=-2.5em,yshift=3.5em]boarder2.center);
	\draw[->, thick] ([xshift=2.5em,yshift=1em]boarder2.center) -- ([xshift=2.5em,yshift=3.5em]boarder2.center);
	\node[] (outputbpe1) at ([xshift=-2.5em,yshift=4.5em]boarder2) {$\hat{X}_{bpe}^{L \times L}$};
	\node[] (outputword1) at ([xshift=2.5em,yshift=4.5em]boarder2) {$\hat{X}_{word}^{L^{'} \times L^{'}}$};

    \end{tikzpicture}
	     }
	     \hspace{.15in}
	    \subfigure[TNT block] { \label{fig:tntblock} 
	    \centering
	    \begin{tikzpicture}[scale=1.3]
	    
	\node[boardernode, fill=goldenorange!30] (boarder) at (-3em,1.5em){};
	
	\node[modulenode, align=center, fill=ugreen] (module1)at (0,0){Inner \\ T-block};
	\node[modulenode, align=center, fill=tiffanyblue] (module2)at ([xshift=-6em,yshift=3em]module1){Outer \\T-block};
	
	\node[circlenode, align=center, fill=ured] (module3)at ([xshift=-6em]module1){+};
	\node[USnode, align=center,rotate=90, font=\small, fill=bangdiblue] (module4)at ([xshift=-3.7em]module1){US};
	
	\draw[->,thick] (module1) edge (module4);
	\draw[->,thick] (module4) edge (module3);
	\draw[->,thick] (module3) edge (module2);

	\node[] (inputbpe) at ([yshift=-4em]module3) {$X_{bpe}^{L \times L}$};
	\node[] (inputword) at ([yshift=-4em]module1) {$X_{word}^{L^{'} \times L^{'}}$};
	\node[] (outputbpe) at ([yshift=4em]module2) {$\hat{X}_{bpe}^{L \times L}$};
	\node[] (outputword) at ([xshift=6em,yshift=4em]module2) {$\hat{X}_{word}^{L^{'} \times L^{'}}$};
	\draw[->,thick] (inputbpe) edge (module3);
	\draw[->,thick] (inputword) edge (module1);
	
	\draw[->,thick] (module2) edge (outputbpe);
	\draw[->,thick] (module1) edge (outputword);

	\end{tikzpicture}
	    
    }

    \caption{The architecture of the re-implementation for TNT.}\label{fig:TNT}
\end{figure*}
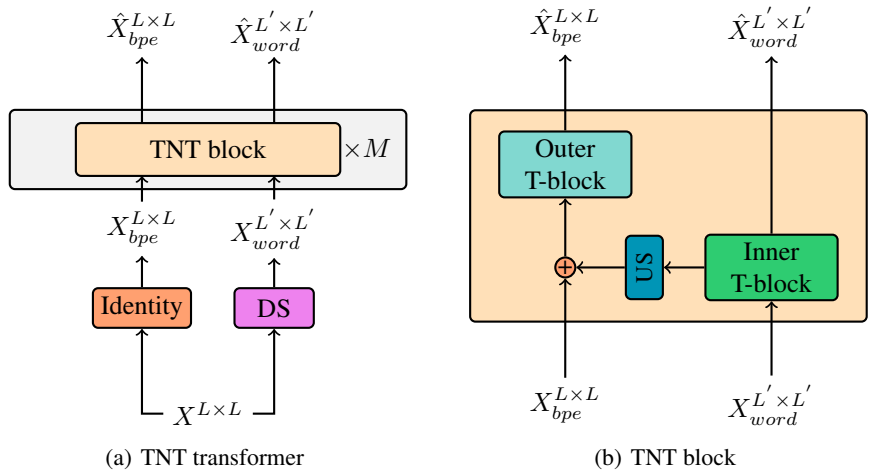
\paragraph{TNT}
\definecolor{upurple}{RGB}{155,89,182}
\definecolor{ublue}{RGB}{52,152,219}
\definecolor{ured}{RGB}{255,160,112}
\definecolor{udark}{RGB}{77,153,77}
\definecolor{ugreen}{RGB}{46,204,113}
\definecolor{darkgreen}{RGB}{61,145,64}
\definecolor{goldenorange}{RGB}{255,153,18}
\definecolor{oldorange}{RGB}{255,153,0}
\definecolor{violet}{RGB}{238, 130,238}
\definecolor{royalblue}{RGB}{65,102,225}
\definecolor{CYAN}{RGB}{65,102,225}

\definecolor{tiffanyblue}{RGB}{129,216,208}
\definecolor{bangdiblue}{RGB}{0,149,182}
\definecolor{kleinblue}{RGB}{0,47,167}
\definecolor{kabuliblue}{RGB}{26,85,153}

	\tikzstyle{circlenode}=[draw,circle,minimum size=4pt,inner sep=0,thick]
    \tikzstyle{add}=[circle,minimum size=1.2em,inner sep=0]
    \tikzstyle{boardernode}=[draw,minimum height=8em,minimum width=15em,inner sep=0,thick,rounded corners=.2em]
    \tikzstyle{boardernode1}=[draw,minimum height=3em,minimum width=15em,inner sep=0,thick,rounded corners=.2em]
    \tikzstyle{boardernode2}=[draw,minimum height=2em,minimum width=10em,inner sep=0,thick,rounded corners=.2em]
    \tikzstyle{modulenode}=[draw,minimum height=2.4em,minimum width=5em,inner sep=0.3em,thick,rounded corners=.2em,thick]
    \tikzstyle{modulenode1}=[draw,minimum height=1.5em,minimum width=3em,inner sep=0.3em,thick,rounded corners=.2em,thick]
    \tikzstyle{USnode}=[draw,minimum height=1.4em,minimum width=5em,inner sep=0em,thick,rounded corners=.2em,thick]

    \tikzstyle{layer} = [draw,minimum height=1.4em,minimum width=15em,inner sep=0,thick,rounded corners=.2em]
    \tikzstyle{stage} = [draw,minimum height=1.4em,minimum width=8em,inner sep=0,thick,rounded corners=.2em]

\begin{figure}[]
  \centering
  \begin{tikzpicture}[scale=0.9]
    \node[] (input) at (0,1em) {$X$};   
    \node[left] (shape1) at (0,2.5em) {$L \times L$};   
    \node[stage,align=center, fill=goldenorange!30] (stage1) at (0em,4em) {Transformer block};   
    \node[] (text1) at ([xshift=6em]stage1) {$\times M_b$};   
    \node[] (text11) at ([xshift=-8em]stage1) {\textbf{Stage 1}};   
    \node[left] (shape2) at (0,5.5em) {$L \times L$};   
    \node[USnode,align=center,fill=purple!50] (DS) at (0em,7em) {DS};   
    \node[left] (shape3) at (0,8.5em) {$L^{'} \times L^{'}$};   
    \node[stage, align=center, fill=goldenorange!30] (stage2) at (0em,10em) {Transformer block};   
    \node[] (text2) at ([xshift=6em]stage2) {$\times M_w$};   
    \node[] (text21) at ([xshift=-8em]stage2) {\textbf{Stage 2}};   
    \node[left] (shape4) at (0,11.5em) {$L^{'} \times L^{'}$};   
    \node[USnode, align=center,fill=purple!50] (gcn) at (0em,13em) {GCN};   %
    \node[left] (shape5) at (0,14.5em) {$L^{'} \times L^{'}$};   
    \node[stage, align=center, fill=goldenorange!30] (stage3) at (0em,16em) {Transformer block};   
    \node[left] (shape6) at (0,17.5em) {$L^{'} \times L^{'}$};   
    
    \node[] (text3) at ([xshift=6em]stage3) {$\times M_p$};   %
    
    \node[] (text31) at ([xshift=-8em]stage3) {\textbf{Stage 3}};   
    \draw[thick, ->] (stage1) edge (DS);
    \draw[thick, ->] (DS) edge (stage2);
    \draw[thick, ->] (stage2) edge (gcn);
    \draw[thick, ->] (gcn) edge (stage3);
    \draw[thick, ->] (input) edge (stage1);
    \draw[thick, ->] ([yshift=0.7em]stage3.center) -- ([yshift=2.5em]stage3.center);

\end{tikzpicture}

  \caption{The architecture of Multi-stage Transformer.}
  \label{fig:multi_stage}
  
\end{figure}
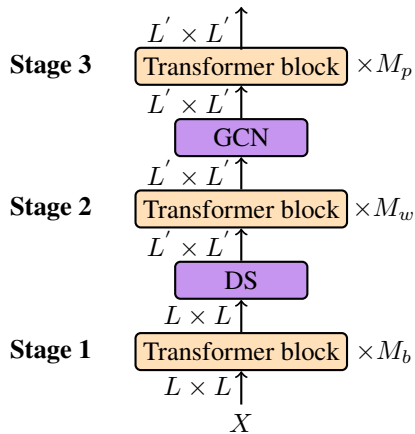
TNT is an improved variant of Vision Transformer \cite{dosovitskiy2020ViT}. It regards the local patches (e.g., 16 $\times$ 16) as ``visual sentences'' and they further split them into much smaller patches (e.g., 4 $\times$ 4) as ``visual words''. There are two data flows in which one flow operates across the visual sentences and the other processes the visual words inside each sentence. Concretely, an inner transformer lock is used to extract correlations among visual words and the output is then added with the original visual sentences representations. At last, an outer transformer block computes the correlations among visual sentences. Also they employ residual connections among inner Transformer blocks to strengthen the information flow. Following this idea, we redesign the Transformer architecture and overall architecture is depicted in Figure \ref{fig:TNT}. Our intention is to extract the inter-group and inter-individual interactions via transformer blocks and outer transformer blocks, respectively.
 
\paragraph{Multi-stage Transformer}
Also, for a more comprehensive comparison, we show the architecture follows a multi-stage fashion. As is shown in Figure \ref{fig:multi_stage}, instead of keeping the length unchanged among different blocks, we partition multi-stages according to the scale. The lower-level is modeling correlations among the individual, which is a mixture sequence of words and sub-words. Subsequently, we used a down-sampling operation to convert individuals into group-level representations. After several stacked transformer blocks, a GCN is employed to embed the phrase-level dependency among words. For a fair comparison, $M_b$, $M_w$ and $M_p$ are set to 2.

\paragraph{Results}

\begin{wraptable}{r}{8.3cm}
      \vspace{-0.3in}
      \caption{Comparisons of \textsc{Umst} with two multiscale variants.}
      \label{tab:comparasion_appendix}
      \vskip 0.2in
      \setlength{\tabcolsep}{4.5pt}
      \small
      \centering
	\begin{tabular}{lrc}
	\toprule
        \bf Model  & \bf Param & \bf BLEU  \\
        \midrule
        Transformer                                      &65M & 27.63   \\
        TNT \cite{han2021transformer}                    &83M & 28.48   \\
        TNT (shared inner and outer)                     &65M & 28.10   \\
        Multi-stage Transformer                          &65M & 27.96   \\
        \textsc{Umst}                                    &70M & 28.51    \\
        \bottomrule
	\end{tabular}
\end{wraptable}


Table \ref{tab:comparasion_appendix} lists the results of \textsc{Umst} and the other two multiscale variants. As we discussed in \Cref{sec:mt}, TNT delivers a BLEU point of 28.48, which outperforms Transformer by a 0.85 BLEU point. To further validate whether the improvement comes from the additional parameters, we also share the parameters between the inner and outer blocks. As we can see, that it can still outperform the baseline by a BLEU point of 0.47. This observation further demonstrates the effectiveness of intra-group interactions. On the other hand, we find that Multi-stage Transformer behaves inferior to our default strategy. A potential explanation is the compression rate of a text is much less than that of an image. 

\subsection{Quantitative Examples on Summarization}
\begin{figure*}[ht]
	\centering
	\subfigure[Standard Transformer]{
	   \label{fig:attn_map_standard_abstract}

       \begin{minipage}[t]{0.5\linewidth}
        \includegraphics[width=1\linewidth]{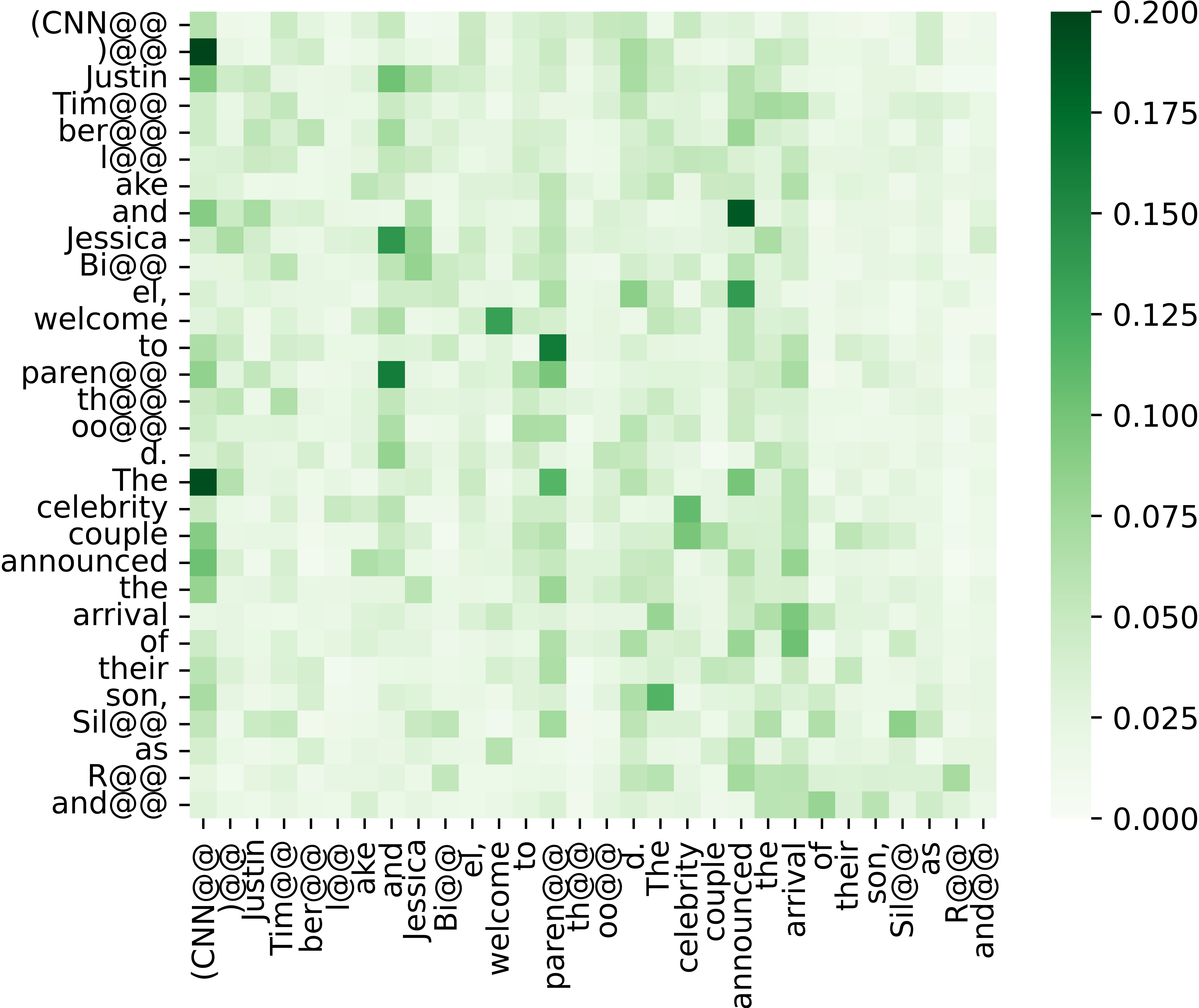}
        \end{minipage}
        \begin{minipage}[t]{0.5\textwidth}
        \includegraphics[width=1\linewidth]{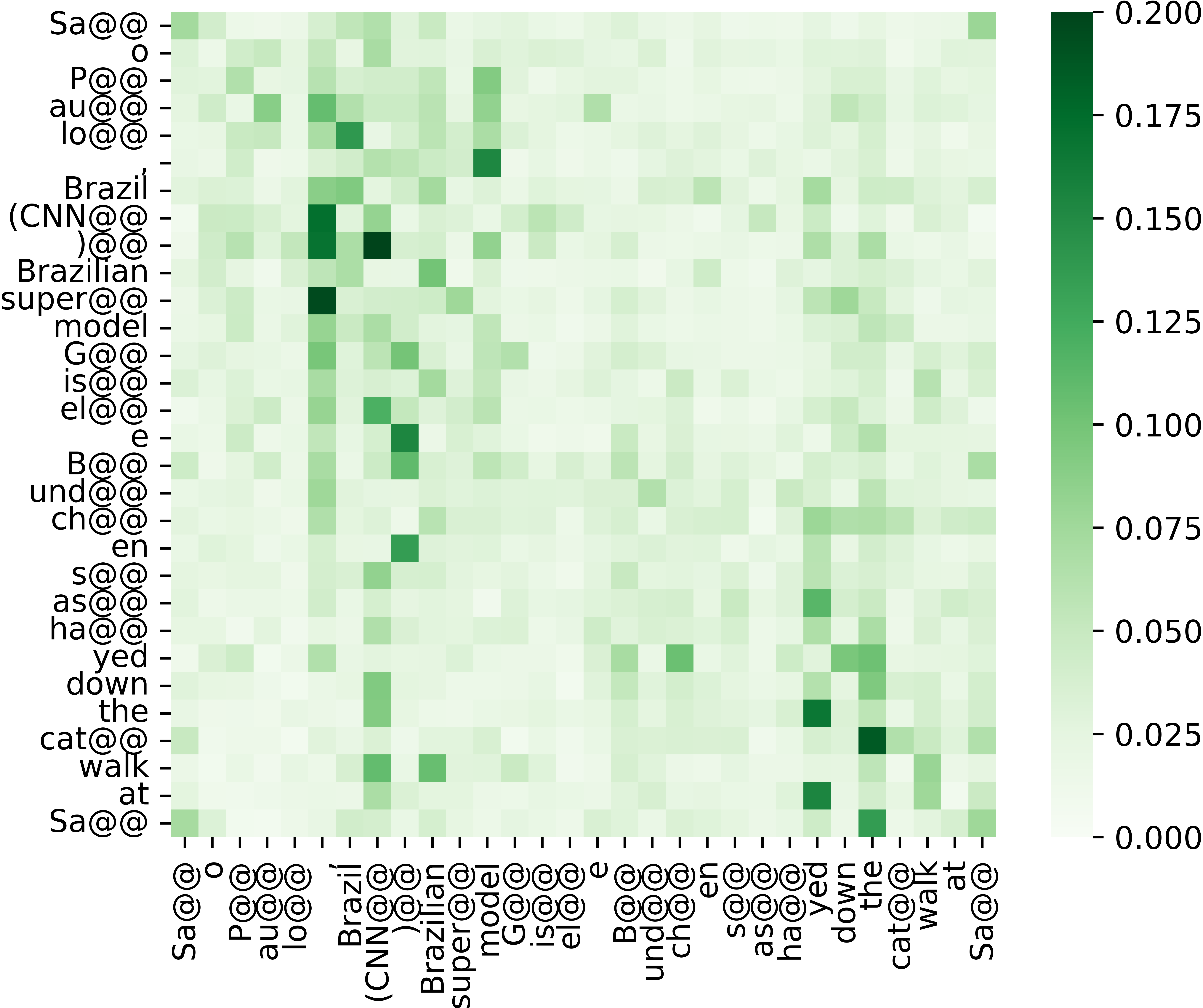}
        \end{minipage}

	}%
	\quad
    \subfigure[Our \textsc{Umst}]{
	   \label{fig:attn_map_our_abstract}

       \begin{minipage}[t]{0.5\linewidth}
        \includegraphics[width=1\linewidth]{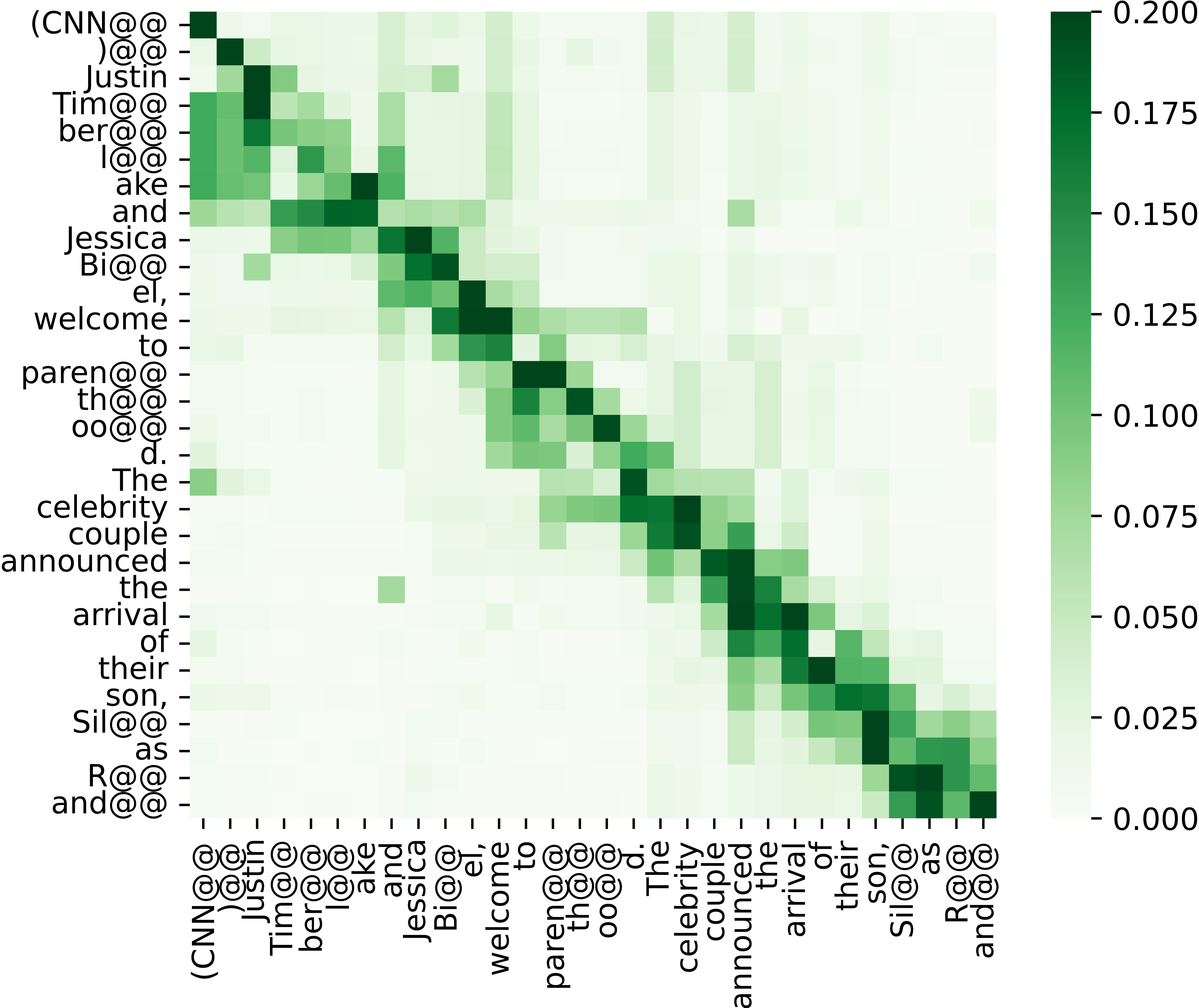}
        \end{minipage}
        \begin{minipage}[t]{0.5\textwidth}
        \includegraphics[width=1\linewidth]{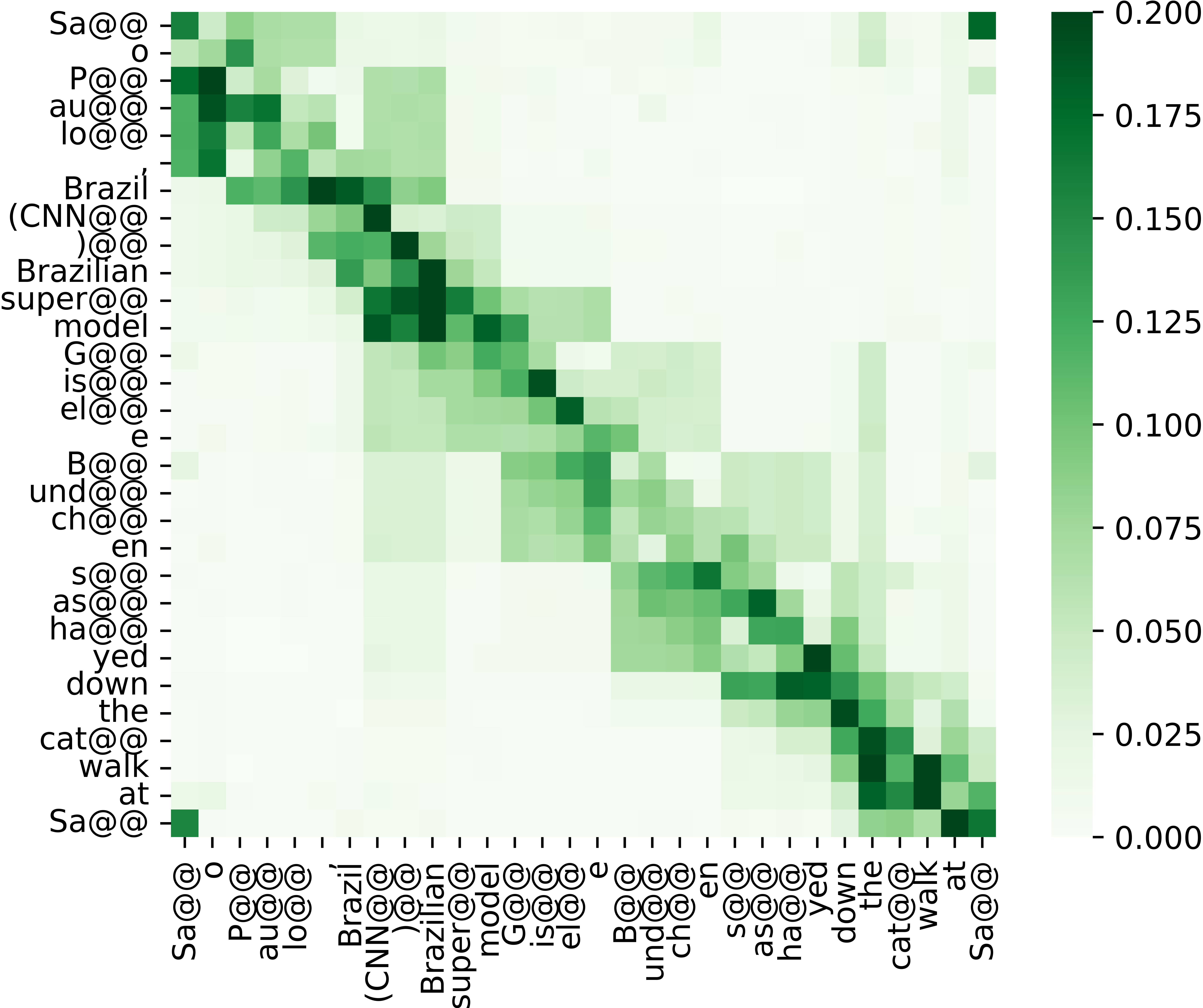}
        \end{minipage}

	}%
	\caption{Quantitative examples of the attention distribution over several cased on abstractive summarization task. The above is the distribution generated by the standard Transformer, and the below is ours (\textsc{Umst}). Dark color means a higher value in the distribution.}
	\label{fig:attn_map_summarization}
\end{figure*}

We have shown the quantitative examples on the machine translation task. Here, we choose a much longer sentence case on the abstractive summarization task. To make a clear presentation, we take parts from a long sentence and re-normalize the attention distribution. Through Figure \ref{fig:attn_map_summarization}, we find similar phenomenon with the machine translation task (see Figure \ref{fig:attn_map}), that sub-words belong to the same word have stronger relationship than distinct individuals. Also the use of inter-group interaction helps the model capture long-range dependencies. However, the attention map of the standard Transformer is mussy and much smoother. The finding here also convinces that multiscale Transformer is indeed helpful.


\end{document}